\documentclass{article} 
\usepackage{iclr2020_conference,times}
\usepackage{CJKutf8}
\usepackage{hyperref}

\hypersetup{
    colorlinks,
    citecolor=black,
    filecolor=black,
    linkcolor=black,
    urlcolor=black,
    linktoc=all
}

\usepackage{bookmark}
\usepackage{url}
\usepackage{graphicx}
\usepackage{amsfonts}
\usepackage{amsmath}
\usepackage{booktabs}
\usepackage{listings}
\usepackage{amsmath}
\usepackage{bbm}
\usepackage{xcolor}
\usepackage{enumitem}
\usepackage{booktabs}
\usepackage[most]{tcolorbox}
\usepackage{color, colortbl}
\usepackage{float}
\usepackage{multirow}
\usepackage{listings}
\usepackage{makecell}
\usepackage{enumitem}

\UseRawInputEncoding

\definecolor{codegreen}{rgb}{0,0.5,0}
\definecolor{codeblue}{rgb}{0.25,0.5,0.5}
\definecolor{codegray}{rgb}{0.6,0.6,0.6}

\usepackage[titles]{tocloft}
\title{ST-MoE: Designing Stable and Transferable Sparse Expert Models}

\author{
Barret Zoph\thanks{Equal contribution. Correspondence to \{\texttt{barretzoph,liamfedus\}@google.com}.} \\
Google Brain \\
\And
Irwan Bello\footnotemark[1] $ $ \thanks{Work was done while at Google.} \\
Google Brain \\
\AND
Sameer Kumar \\
Google \\
\And
Nan Du \\
Google Brain \\
\And
Yanping Huang \\
Google Brain \\
\AND
Jeff Dean \\
Google Research \\
\And
Noam Shazeer\footnotemark[2] \\
Google Brain \\
\And
William Fedus\footnotemark[1] \\
Google Brain \\
}

\iclrfinalcopy 

\begin{document}
\maketitle

\begin{abstract}
Scale has opened new frontiers in natural language processing -- but at a high cost.
In response, Mixture-of-Experts (MoE) and Switch Transformers have been proposed as an energy efficient path to even larger and more capable language models.
But advancing the state-of-the-art across a broad set of natural language tasks has been hindered by training instabilities and uncertain quality during fine-tuning.
Our work focuses on these issues and acts as a design guide.
We conclude by scaling a sparse model to 269B parameters, with a computational cost comparable to a 32B dense encoder-decoder Transformer (Stable and Transferable Mixture-of-Experts or ST-MoE-32B).
For the first time, a sparse model achieves state-of-the-art performance in transfer learning, across a diverse set of tasks including reasoning (SuperGLUE, ARC Easy, ARC Challenge), summarization (XSum, CNN-DM), closed book question answering (WebQA, Natural Questions), and adversarially constructed tasks (Winogrande, ANLI R3).
\footnote{Code for our models is available at \url{https://github.com/tensorflow/mesh/blob/master/mesh_tensorflow/transformer/moe.py}}
\end{abstract}

\newpage
\setcounter{tocdepth}{2} 
\tableofcontents
\newpage

\section{Introduction}

Sparse expert neural networks showcase the advantage of sheer scale and offer an efficient alternative to the static neural network architectures commonly used today \citep{raffel2019exploring,brown2020language,rae2021scaling}.
Rather than applying the same parameters to all inputs, sparse expert networks dynamically select which parameters to use for each input \citep{shazeer2017outrageously}. 
This allows for networks to vastly expand their number of parameters, while keeping the FLOPs per token roughly constant. 
These approaches have yielded state-of-the-art translation models \citep{lepikhin2020gshard}, 4-7x pre-training speed-ups \citep{fedus2021switch,artetxe2021efficient}, and GPT-3 level one-shot performance using 1/3 the energy training cost \citep{du2021glam}.
And despite a shocking number of parameters, sparse models reduce the carbon footprint for training large neural networks by an order of magnitude \citep{patterson2021carbon}.
However, difficulties remain.

\cite{fedus2021switch} observed that a sparse 1.6T parameter model achieved a 4x pre-training speed-up over the prior state-of-the-art \citep{raffel2019exploring}, but lagged smaller models when fine-tuned on common benchmarks like SuperGLUE.
Similar gaps were observed in \cite{artetxe2021efficient} when MoE language models were fine-tuned on out-of-domain data.
In response, Switch-XXL, a model with fewer parameters, but a 8x-larger computational footprint (FLOPs approximately equal to the largest T5 model), was proposed and improved quality on natural language understanding tasks.
However, necessary pre-training was hampered by training instabilities previously undetected during smaller scale studies.
These instabilities were later identified in other sparse models \citep{du2021glam}.
These results revealed a necessary balance of parameters \emph{and} computation, but left an open question on how to reliably train these types of models.

Our aim in this paper is to increase the \emph{practicality} and \emph{reliability} of sparse models.
We study these two issues and pre-train a 269B sparse model that achieves state-of-the-art results when fine-tuned across many competitive NLP benchmarks, including SuperGLUE.
We also put forth additional analysis and a design guide (or at least, our heuristics) for sparse expert models.
Furthermore, this work emphasizes \emph{jointly} optimizing both the upstream pre-training and the downstream fine-tuning metrics to avoid discrepancies \citep{tay2021scale}.

\begin{tcolorbox}[enhanced,attach boxed title to top center={yshift=-3mm,yshifttext=-1mm}, title=\textbf{Contributions}, colback=white, colframe=white!75!blue, coltitle=black, colbacktitle=white]
    \begin{enumerate}[leftmargin=*,label=\textbf{\arabic*.}]
    \item A large-scale study of the quality-stability trade-offs of stability techniques.
    \item An introduction of the \emph{router z-loss} that resolves instability issues, while slightly improving model quality.
    \item A fine-tuning analysis of sparse and dense models highlighting different hyperparameter sensitivity to the batch size and learning rate. 
    We show bad hyperparameters result in virtually no fine-tuning gain over dense models, despite large pre-training speed-ups.
    \item Architectural, routing and model design principles for designing Pareto efficient sparse models in a distributed setting.
    \item A qualitative analysis tracing token routing decisions across expert layers.
    \item A 269B sparse model (the Stable Transferable Mixture-of-Experts or ST-MoE-32B) which achieves state-of-the-art performance across a diverse set of natural language benchmarks.
    \end{enumerate}
\end{tcolorbox}

\section{Background}
\label{sec:background}
Sparse expert models typically substitute a neural network layer with a set of \emph{experts}, each having unique weights \citep{jacobs1991adaptive,jordan1994hierarchical}.
Typically all the experts within a layer are of the same type and shape (homogeneous), however, varied (heterogeneous) expert-types are possible.
Inputs are only processed by a subset of the experts to save computation, so a mechanism must be added to determine where to send each input. 
Usually a \emph{router} or gating network determines where to send inputs (i.e.\@ words, sentences, image patches, etc.), but alternative schemes have been proposed \citep{lewis2021base,roller2021hash,zuo2021taming,clark2022unified}.

Specifically, in natural language processing, \citet{shazeer2017outrageously} proposed a Mixture-of-Experts (MoE) layer which takes a token representation $x$ as input and routes it to the best matched top-$k$ experts selected out of a set $\{E_i(x)\}_{i=1}^N$ of $N$ experts.
The router variable $W_r$ produces logits $h(x) = W_r \cdot x$ which are normalized via a softmax distribution over the available $N$ experts at that layer.
The gate-value for expert $i$ is given by
\begin{equation}
    p_i(x) = \frac{e^{h(x)_i}}{\sum_j^N e^{h(x)_j}}
\end{equation}
and the token $x$ is routed to the experts with the highest top-$k$ gate values (set of indices $\mathcal{T}$).
The output of the layer is the weighted sum of each expert's computation by the gate value
\begin{equation}
    y = \sum_{i \in \mathcal{T}} p_i(x) E_i(x) 
\end{equation}
Originally proposed in LSTMs \citep{hochreiter1997long}, expert layers were later used in the Transformer \citep{vaswani2017attention} by \cite{shazeer2018mesh} and \cite{lepikhin2020gshard}.
Follow-on work by \cite{fedus2021switch} simplified the MoE further to route tokens to a single expert (top-1) and reduced other costs to improve training efficiency.

To improve hardware utilization, most implementations of sparse models have static batch sizes for each expert~\citep{shazeer2017outrageously,shazeer2018mesh,lepikhin2020gshard,fedus2021switch}.
The \emph{expert capacity} refers to the number of tokens that can be routed to each expert.
If this capacity is exceeded (the router sends too many inputs to that expert) then the overflowed tokens have no computation applied to them and are passed to the next layer through a residual connection.

\begin{table}[!ht]
    \centering
    \small
    \begin{tabular}{l|p{10cm}}
        \toprule
        \textbf{Terminology} & \textbf{Definition}  \\
        \midrule
        \textbf{Expert} & An independently-learned neural network with unique weights.\\
        \midrule
        \textbf{Router} & A network that computes the probability of each token getting sent to each expert.\\
        \midrule
        \textbf{Top-$n$ Routing} & Routing algorithm where each token is routed to $n$ experts. \\
        \midrule
        \textbf{Load Balancing Loss} & An auxiliary (aux) loss to encourage each group of tokens to evenly distribute across experts.\\
        \midrule
        \textbf{Group Size} & The global batch size is split into smaller groups, each of size Group Size. Each group is considered separately for load balancing across experts. Increasing it increases memory, computation, and communication.\\
        \midrule
        \textbf{Capacity Factor (CF)} & Each expert can only process up to a fixed number of tokens, which is often set by evenly dividing across experts, $\frac{\text{tokens}}{\text{experts}}$. The capacity factor can expand or contract this amount to $\text{CF} \cdot \frac{\text{tokens}}{\text{experts}}$. \\
        \midrule
        \textbf{FFN} & Acronym of Feed Forward Network (FFN) layer of Transformer consisting of linear, activation, linear. \\
        \midrule
        \textbf{Encoder-Decoder} & A Transformer architectural variant that all of our models are based on. Consists of an encoder that does all-to-all attention on the inputs and a decoder that attends to the encoder and to its own inputs in an autoregressive manner. \\
        \midrule
        \texttt{allreduce} & Communication primitive which sums a subset of $n$ tensors on $n$ different devices, then broadcasts the summed value to all $n$ devices. This is used in distributed training for gradient accumulation and model parallelism.\\
        \midrule
        \texttt{all2all} & Communication primitive where each device sends to every other device a part of its tensor. Used in sparse Transformer models for token routing.\\
        \midrule
        ($\uparrow$/$\downarrow$) & Indicates whether higher/lower values are better (e.g.\@ accuracy/train loss).\\
        \bottomrule
    \end{tabular}
    \caption{\textbf{Terminology used throughout the paper.}}
    \label{tab:terminology}
\end{table}

The batch $\mathcal{B}$ of input tokens is broken into $\mathcal{G}$ unique groups across the data-parallelism dimension\footnote{Our implementation relies on einsums with one-hot tensors for dispatching and combining tensors to/from experts. The size of this one-hot tensor grows quadratically with the number of tokens being routed as a group which motivates breaking the batch into smaller groups. This may be avoided with sparse lookup operations.}, each with size $\mathcal{B}/\mathcal{G}$.
The expert capacity is equal to $ \text{CF} \cdot\text{tokens} / \text{experts}$ where $\text{CF}$ represents the capacity factor hyperparameter, $\text{experts}$ is the number of experts and $\text{tokens}$ is the group size. 
If the capacity factor is increased, it creates extra buffer so that fewer tokens will be dropped in case of load imbalance. However, increasing the capacity factor also increases the memory and computational costs, so there exists a trade off\footnote{See \cite{fedus2021switch} for a graphical illustration of how the capacity factor works.}.

Finally, an auxiliary load balancing loss encourages tokens to be roughly evenly distributed across the experts \citep{shazeer2017outrageously}.
This improves the hardware efficiency by ensuring that all accelerators are processing significant chunks of data in parallel as mentioned above. 
The details of the loss are presented in Appendix \ref{appendix:load_balance}.
However, alternatives exist: \cite{lewis2021base} and \cite{clark2022unified} treats balanced token allocation as an assignment problem and removes the auxiliary loss entirely.

\section{Stabilizing Training of Sparse Models}
\label{sec:stability}

\begin{figure}[ht!]
    \centering
    \includegraphics[width=0.45\columnwidth]{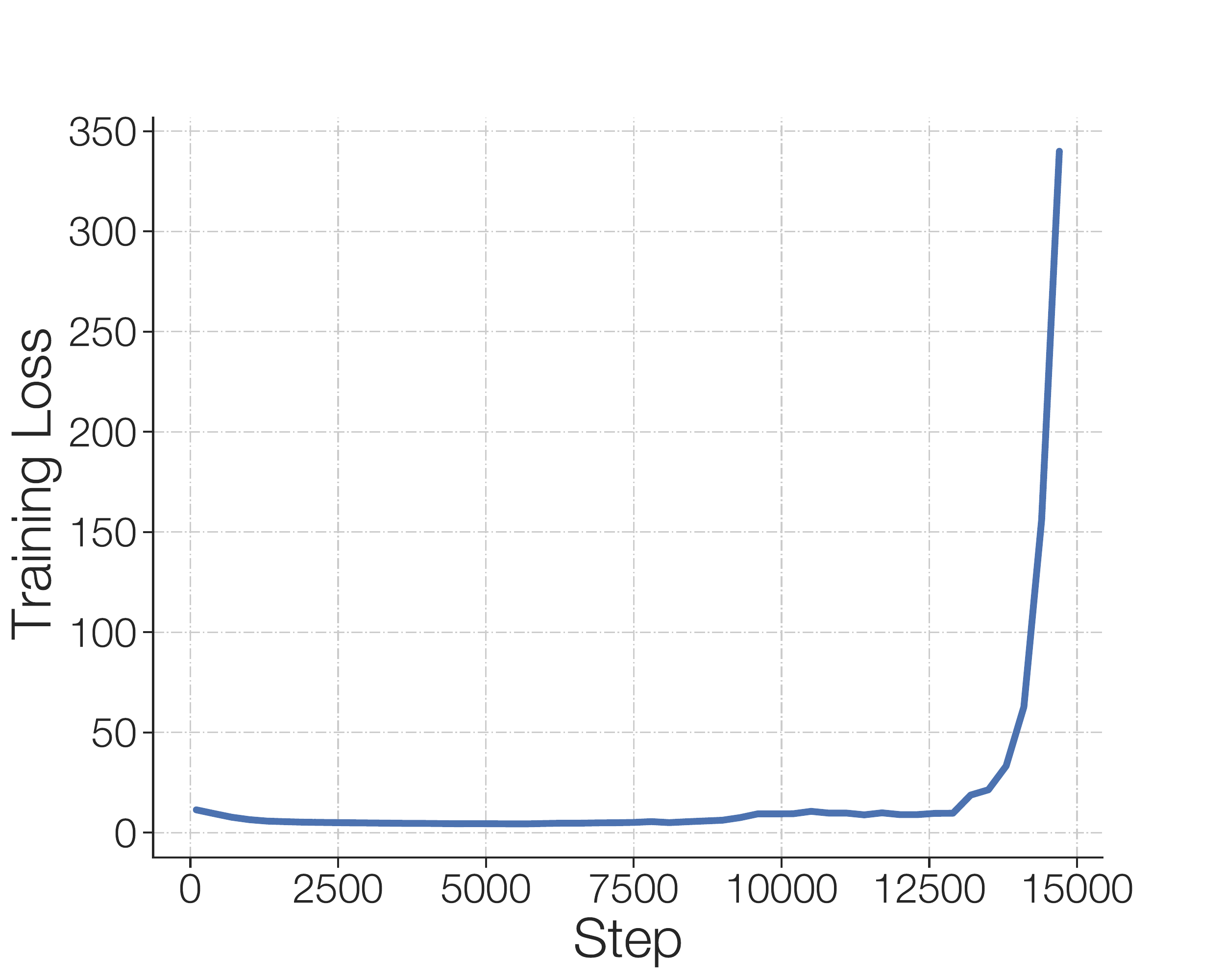}
    \includegraphics[width=0.45\columnwidth]{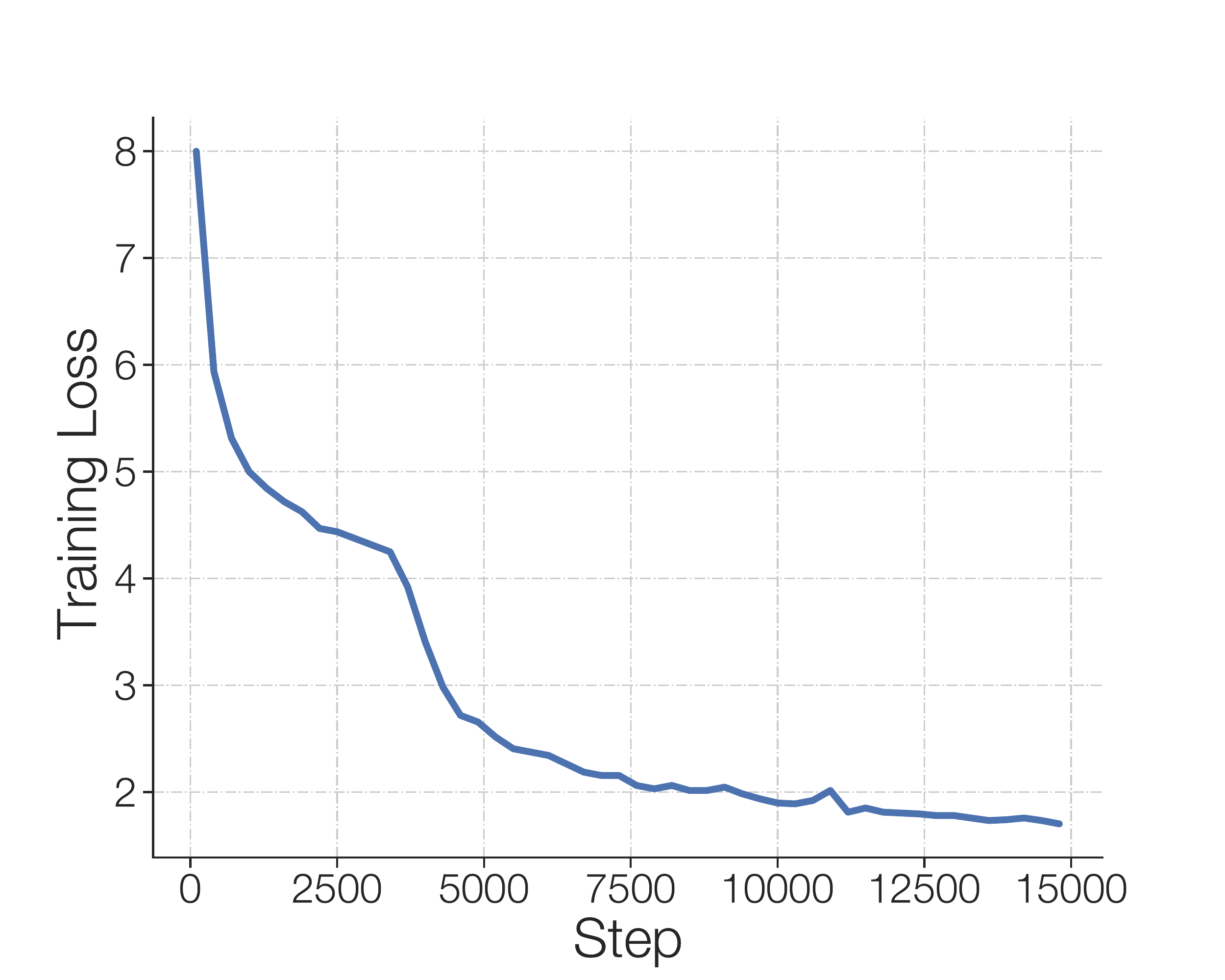}
    \caption{\textbf{Training instabilities for sparse models.} We refer to training instabilities as divergences in the training loss. Above are two runs from sparse models FLOP-matched to the T5-XL version \citep{raffel2019exploring} each trained with a batch size of 1M tokens using the Adafactor optimizer \citep{shazeer2018adafactor}.
    \textbf{(Left)} An unstable training run. \textbf{(Right)} A stable training run.
    }
    \label{fig:instability}
\end{figure}

Sparse models often suffer from training instabilities (Figure \ref{fig:instability}) worse than those observed in standard densely-activated Transformers.  

It's straightforward to find changes that improve the \emph{stability}, however, these often come at an untenable expense to model \emph{quality} (for instance, using an arbitrarily small learning rate or using tight gradient clipping). 
We categorize and examine several approaches to improve stability.
The stability techniques span generic fixes to Transformers as well as those specific to sparse models:
\textbf{(1)} Remove multiplicative interactions \textbf{(2)} Inject model noise \textbf{(3)} Constrain activations, and gradients.
We conclude with our recommendation: a new auxiliary loss, the \emph{router z-loss}, which significantly improves training stability with no quality degradation.
This is an adaptation of the z-loss used for final softmax logits in the Mesh Tensorflow codebase \citep{shazeer2018mesh}.

\begin{tcolorbox}[enhanced,attach boxed title to top center={yshift=-3mm,yshifttext=-1mm}, title=\textbf{Stabilizing Sparse Models}, colback=white, colframe=white!75!blue, coltitle=black, colbacktitle=white]
    \begin{enumerate}[leftmargin=*,label=\textbf{\arabic*.}]
        \item Many methods stabilize sparse models, but at the expense of worse quality.
        \item The router z-loss stabilizes models without quality degradation.
        \item Transformer modifications with more multiplicative components (GEGLU, RMS normalization) worsen stability, but boost quality.
    \end{enumerate}
\end{tcolorbox}

\paragraph{Designing a large-scale stability study.} We design a large-scale stability study of sparse models FLOP-matched to the T5-XL version \citep{raffel2019exploring} pre-trained on the multilingual corpus mC4 \citep{xue2020mt5}.
Each sparse model has 32 experts and we introduce a sparse MoE layer for every fourth FFN. The train capacity factor is 1.25 and the eval capacity factor is 2.0. See Table \ref{tab: model_params} for a more detailed description of models used throughout this paper.
For each stability technique, we record the fraction that are stable, the mean quality (negative log perplexity on English), and the standard deviation over seeds.

The primary issue in constructing this study is that small models are rarely unstable but large unstable models are too costly to run for sufficient steps and seeds.
We found a sparse model FLOP-matched to T5-XL to be good object of study because it was unstable roughly $1/3$ of the runs, but was still relatively cheap to train. 
Furthermore, we run our instability experiments on multilingual data since we find this exacerbates model instabilities, allowing us to experiment on slightly smaller models. 
See Section~\ref{sec:discussion} for more details.
Our baseline configuration is trained using six random seeds and each configuration with a stability technique uses three random seeds.
We use six seeds for the baseline to better characterize the instability rate and three seeds for the variants to save compute.
Each model is pre-trained for 20k steps on mC4 using a masked language modeling objective \citep{fedus2018maskgan,devlin2018bert}.

\subsection{Stability and Quality Tradeoffs when Removing Multiplicative Interactions}
\label{sec:stability_mult}

Some architectural improvements involve more multiplications than additions or do not sum many items at once.
For example, a matrix multiplication has one multiplication for each addition and
hence we do not refer to it as a ``multiplicative" operation. 
We present and analyze the impact of two instances of multiplicative interactions in Transformers here.

\paragraph{GELU Gated Linear Units (GEGLU).} Our first example is the Gated Linear Unit \citep{dauphin2017language} which is a component-wise product of two linear projections, one of which is first passed through a sigmoid function. 
\cite{shazeer2020glu} extends this to other variants and presents a GELU-Linear \citep{hendrycks2016gaussian} FFN layer as a replacement the usual ReLU \citep{nair2010rectified} FFN in Transformer. 
\begin{equation}
    FFN_{GEGLU}(x, W, V, b, c) = GELU(xW + b) \odot (xV + c)
\end{equation}
This quality gain was corroborated in later work \citep{narang2021transformer}.

\paragraph{Root Mean Square Scale Parameters.} Our second example is the scale parameter in root mean square (RMS) normalization \citep{zhang2019root}. 
Within the Transformer, rather than calling layers back-to-back, there is an internal structure (referred to as sublayer calls) which improve gradient propagation and training dynamics.
Our sublayer calls match that of \cite{raffel2019exploring} and consist of: 
\textbf{(1)} RMS normalization, \textbf{(2)} layer call (e.g. Self Attention), \textbf{(3)} dropout \citep{srivastava2014dropout}, \textbf{(4)} add residual \citep{he2015deep}.
RMS normalization scales the input vector $x \in \mathbb{R}^d$ element-wise per the root-mean-square. It then rescales the output element-wise by multiplying with a learned scale parameter $g$.
\begin{equation}
y_i =  \frac{x_i}{\sqrt{ \frac{1}{d} \sum_{i=1}^d x_i^2}} \cdot g_i  
\end{equation}

Table \ref{tab:stability_multiplicative} shows that both removing GEGLU layers or the RMS scale parameter improves stability, but at a significant loss to model quality.
We note that these scale parameters ($g$) have a disproportionate gain to model quality versus parameters elsewhere (e.g. FFN).  
In line with our findings, \cite{shleifer2021normformer} found adding a learned multiplicative scalar to the residual connection in Transformers made them much more unstable. 

\begin{table}[!th]
\centering
    \begin{tabular}{ccc}
    \toprule
        \textbf{Method} & \textbf{Fraction Stable} & \textbf{Quality ($\uparrow$)} \\ 
        \midrule
        Baseline & $4/6$ & \textbf{-1.755} $\pm 0.02$\\ 
        Remove GEGLU & $3/3$ & -1.849 $\pm 0.02$ \\
        Remove RMS Norm. Scale Param & $3/3$ & -2.020 $\pm 0.06$ \\
    \bottomrule
    \end{tabular}
    \caption{\textbf{Removing operations with more multiplicative interactions}. Multiplicative interactions improve quality, but can destabilize training. Individually removing two sources of multiplicative components improves the stability, but worsens quality significantly. When we remove the GEGLU layer, we replace it with with an equivalent Dense-ReLU-Dense layer to match the FLOPs and parameters.}
    \label{tab:stability_multiplicative}
\end{table}

In Appendix~\ref{appendix: architecture}, we further study the quality impact of adding new multiplicative interactions in expert layers.
We find that this operation yields quality improvements with virtually no slow-down in model step time.

\subsection{Stability and Quality Tradeoffs when Adding Noise} 
We next explore a hypothesis that adding noise into the model can improve training stability ~\citep{neelakantan2015adding}. 
\cite{taleb2012antifragile} argues that certain systems exhibit the property of anti-fragility, where they \emph{improve} through noise.
Inspired by the concept and by our observation that fine-tuning (which injects noise via dropout) was rarely unstable, we examined whether training noise might improve the stability of sparse models. 
Table \ref{tab:stability_noise} shows a stability improvement versus the baseline, but at the expense of lower quality.
We also find that input-jitter, introduced by \cite{fedus2021switch}, diminishes quality at XL-scale, hence we ablate it in our models. Input-jitter multiplies the input logits to the router by a uniform random variable between $[1 - 10^{-2}, 1 + 10^{-2}]$ . Dropout in our ablation is applied throughout the Transformer.
As seen previously, improvements in small-scale settings may fail to generalize when scaled up and therefore trends should always be monitored and re-assessed at increasing scale \citep{kaplan2020scaling}.

\begin{table}[!th]
\centering
    \begin{tabular}{ccc}
    \toprule
        \textbf{Method} & \textbf{Fraction Stable} & \textbf{Quality ($\uparrow$)} \\ 
        \midrule
        Baseline & $4/6$ & \textbf{-1.755} $\pm 0.02$\\  
        Input jitter ($10^{-2}$) & $3/3$ & -1.777 $\pm 0.03$ \\
        Dropout (0.1) & $3/3$ & -1.822 $\pm 0.11$\\
    \bottomrule
    \end{tabular}
    \caption{\textbf{Injecting noise during training}. Both input-jitter and dropout improve stability, but lead to a significant loss of model quality. There is a clear tradeoff with most methods: when one improves stability, it then typically decreases model quality. Our work aims to find methods that fix stability without hurting quality. 
    }
    \label{tab:stability_noise}
\end{table}

\subsection{Stability and Quality Tradeoffs when Constraining Activations and Gradients}
\label{sec:z_loss}
One of the most successful approaches to stabilizing neural networks are constraints on activations, and gradients \citep{pascanu2013difficulty,ioffe2015batch,salimans2016weight,ba2016layer}.
A popular approach consists in the clipping of gradient norms to remedy exploding gradients while backpropagating through deep networks \citep{pascanu2013difficulty}.

In this work, we use the Adafactor optimizer due to its memory efficiency (though recently introduced 8-bit optimizers \citep{dettmers20218bit} may offer better trade-offs).
Instead of gradient clipping, Adafactor uses \emph{update clipping}, where the changes to the weights are constrained to be below a certain norm.
We experiment with tightening the update clipping to a smaller value.

Next, we study constraints on the logits going into the router.
The router computes the probability distribution over the experts in \texttt{float32} precision (i.e.\@ selective precision) \citep{fedus2021switch}.
However, at the largest scales, we find this is insufficient to yield reliable training.
To fix this, we introduce the \emph{router z-loss},
\begin{equation}
    L_{z}(x) = \frac{1}{B} \sum_{i=1}^B \left(\log \sum_{j=1}^N e^{x_j^{(i)}} \right)^2
\label{eqn: z_loss}
\end{equation}
where $B$ is the number of tokens, $N$ is the number of experts, and $x \in \mathcal{R}^{B \times N}$ are the logits going into the router. 
This penalizes large logits into the gating network and Section~\ref{sec:prec_format} contains a more detailed explanation of why the z-loss before the router is useful. 

Table \ref{tab:stability_constraint} shows that both update clipping and the router z-loss stabilize the model in all 3 runs, but the update clipping significantly hurts the model quality.
Therefore we use the z-loss method for fixing our model stability due to improved quality and stability\footnote{We also experimented with adding z-losses onto the attention logits which also improves model instability without hurting model quality.}.

\begin{table}[!th]
\centering
    \begin{tabular}{ccc}
    \toprule
        \textbf{Method} & \textbf{Fraction Stable} & \textbf{Quality ($\uparrow$)} \\ 
        \midrule
        Baseline & $4/6$ & -1.755 $\pm 0.02$\\ 
        Update clipping ($\text{clip}=0.1$) & $3/3$ & -4.206 $\pm 0.17$ \\
        Router Z-Loss & $3/3$ & \textbf{-1.741} $\pm 0.02$ \\
    \bottomrule   
    \end{tabular}
    \caption{\textbf{Constraining weight updates and router logits.} Constraining the update clipping in Adafactor improves stability, but at a catastrophic loss of quality. Looser clipping values did not reliably stabilize training so we exclude them here. The router z-loss stabilizes the model without any quality degradation (in this case, we observe a slight quality boost).}
    \label{tab:stability_constraint}
\end{table}

The router z-loss introduces another hyperparameter $(c_{z})$, which is the coefficient to weight this as part of the total loss optimized.
The total loss is a linearly weighted combination of the cross entropy loss ($L_{CE}$), the auxiliary load balance loss ($L_{B}$), and the router z-loss ($L_{Z}$), yielding a total loss 
\begin{equation}
L_{tot} = L_{CE} + c_B L_{B} + c_z L_{Z}   
\end{equation}
We choose a value of $c_z=0.001$ based on the best model quality after pre-training with a hyperparameter sweep.
Appendix \ref{app: router_z_loss} logs the resulting losses over the course of pre-training.

\subsection{Selecting a Precision Format: Trading Efficiency and Stability}
\label{sec:prec_format}

As in most modern distributed Transformers we train with \emph{mixed precision} \citep{micikevicius2017mixed} \footnote{See Mesh Tensorflow for implementation details: \url{https://github.com/tensorflow/mesh/blob/master/mesh_tensorflow/}}. 
Weights are stored in \texttt{float32} for gradient updates and then converted to \texttt{bfloat16} when doing matrix multiplications in the forward and backward pass\footnote{Matrix multiplications on TPUs perform multiplications in \texttt{bfloat16} and accumulations in \texttt{float32}.}.
Furthermore, all activations are stored and operated on in \texttt{bfloat16} and \texttt{allreduce} communications can be done in either \texttt{bfloat16} or \texttt{float32} numerical precision.
For the largest model explored in this work (ST-MoE-32B presented later) we find speed-ups halving the numerical precision of the \texttt{allreduce}, however this also can destabilize the training so we keep this as \texttt{float32} throughout this work.

A lower precision format enables more efficient models by reducing
\textbf{(a)} communication costs between processors and memory,
\textbf{(b)} computation costs,
\textbf{(c)} memory for storing tensors (e.g.\@ activations).
However, lower precision formats come at the expense of larger roundoff errors which can lead to irrecoverable training instabilities.

\begin{figure}[ht!]
    \centering
    \includegraphics[width=0.8\columnwidth]{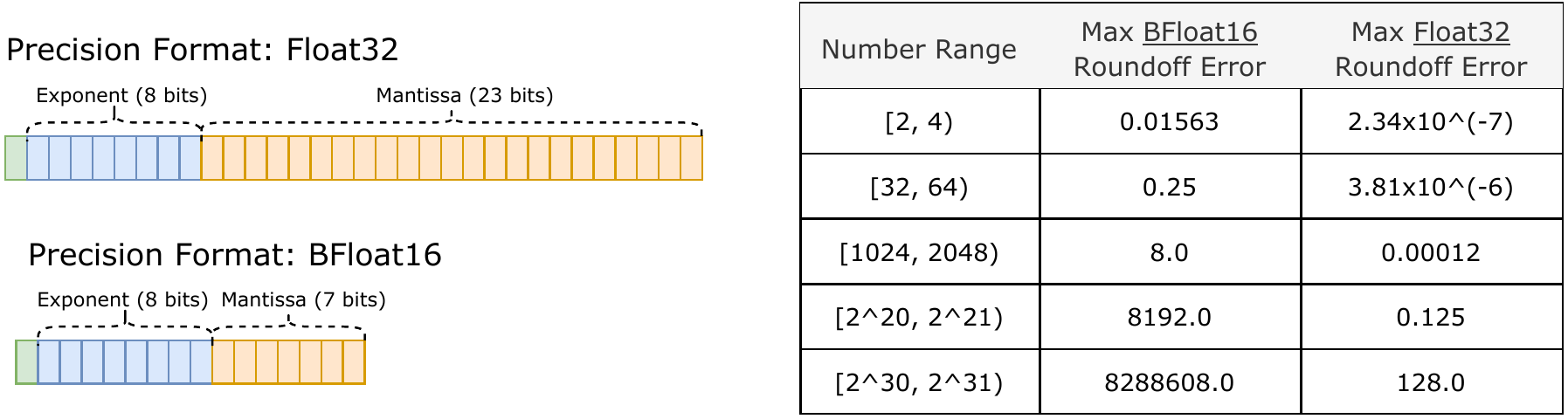}
    \caption{\textbf{Numerical precision formats and roundoff errors.} Larger numbers have larger roundoff errors. \texttt{bfloat16} has up to 65,536x worse roundoff errors than \texttt{float32}. The router z-loss encourages the absolute magnitude of numbers to be small, which doesn't hinder model performance and reduces roundoff errors. The router z-loss is most effective into functions where larger errors can drastically change the relative output (e.g.\@ exponential and sinusoidal functions).}
    \label{fig:float_diagram}
\end{figure}

\paragraph{Understanding precision format and roundoff errors.} 
Figure~\ref{fig:float_diagram} reviews the properties of different precision formats and their corresponding roundoff errors for different number ranges.
Numbers in any range of two consecutive powers of 2 (e.g.\@ [2,4) and [1024, 2048)) are represented by a fixed number of mantissa bits (7 for \texttt{bfloat16}, 23 for \texttt{float32}).
As a result, \textbf{(1)} \texttt{bfloat16} will have about 65,536x (i.e. $23-7=16$ additional bits and $2^{16}=65536$) as large roundoff errors as \texttt{float32} and \textbf{(2)} larger numbers have larger roundoff errors. Due to the 8 exponent bits, number can get as large as $\approx3e^{38}$, which leads to even \texttt{float32} having some issues with roundoff errors.

\paragraph{Sparse expert models are sensitive to roundoff errors because they have more exponential functions due to the routers.}
Sparse expert models introduce additional exponential functions -- through the router -- which can exacerbate roundoff errors\footnote{Exponential functions have the property that a small input perturbation can lead to a large difference in the output. As an example, consider inputting 10 logits to a softmax function with values of 128 and one logit with a value 128.5. A roundoff error of 0.5 in \texttt{bfloat16} will alter the softmax output by 36\% and incorrectly make all logits equal. The calculation goes from $\frac{\exp(0)}{\exp(0) + 10 \cdot \exp(-0.5)}\approx0.142$ to $\frac{\exp(0)}{\exp(0) + 10 \cdot \exp(0)}\approx0.091$. This occurs because the max is subtracted from all logits (for numerical stability) in softmax operations and the roundoff error changes the number from 128.5 to 128. 
This example was in \texttt{bfloat16}, but analogous situations occur in \texttt{float32} with larger logit values.} and lead to training instabilities.
While a roundoff error does not change the ordering of probabilities within a softmax operation, it does impact the routing of the second token in MoE due to relative thresholding (e.g.\@ a token is only routed to its second place expert if the gating probability for the second expert is $1/5$ as large as that of the first expert).
Additionally, roundoff errors can drastically change the probability that scales the expert output -- which we have found to be important. 
Finally, we conjecture that the higher stability we observed for decoder-only models (not shown here) was because they had fewer exponential functions.
Section~\ref{sec:discussion} contains a more detailed discussion. 

\paragraph{An aside on the router z-loss.} 
One might think that the router z-loss is a convoluted method replaceable by clipping logits ~\citep{wu2016google}.
We explain why this is not the case.
The goal is to minimize large roundoff errors going into exponential functions.
Clipping the logits occurs \emph{after} any roundoff errors -- resulting in even larger discontinuities.
In one view, clipping in itself is a roundoff error; conversely, the z-loss naturally encourages the model to produce logits that are small in value and thus more accurately modeled.
Due to these dynamics, we ensure all exponentiated tensors are cast to \texttt{float32}.
This hints at the possibility of better number formats for neural networks because of the unused exponent bits when z-losses are added throughout the network (see Section~\ref{sec:discussion}).

\section{Fine-Tuning Performance of Sparse Models}
The best performing language models are usually obtained by (1) pre-training on large amounts of data (e.g.\@ the internet) followed by (2) fine-tuning on a task of interest (e.g.\@ SuperGLUE). 
Promising new techniques have emerged as an alternative, including few-shot inference \citep{brown2020language}, prefix tuning \citep{li2021prefix}, prompt tuning \citep{lester2021power}, and adapter modules \citep{houlsby2019parameter}  -- however, a quality gap still persists compared to fine-tuning.
Because of this, we focus on fine-tuning in this work, but highlight recent successes of sparse models in few-shot settings from \cite{du2021glam,artetxe2021efficient}.
Further, we leave as future work techniques that adapt large language models through reinforcement learning \citep{ouyang2022training}

\subsection{Hypothesis: A Generalization Problem}
\label{sec: overfitting}
Sparse models have performed remarkably well in the regime of large datasets, but have sometimes performed poorly when fine-tuning \citep{fedus2021switch,artetxe2021efficient}.
We present evidence for a (not so surprising) hypothesis that sparse models are prone to overfitting.
We illustrate this problem through two tasks in SuperGLUE \citep{wang2019superglue} -- Commitment Bank \citep{de2019commitmentbank} and ReCORD \citep{zhang2018record}.
Commitment Bank (CB) has 250 training examples while ReCORD has over 100,000.
This significant size discrepancy facilitates a natural study for overfitting on two tasks selected as part of the same benchmark.

\begin{figure}[h!]
    \centering
    \includegraphics[width=0.49\columnwidth]{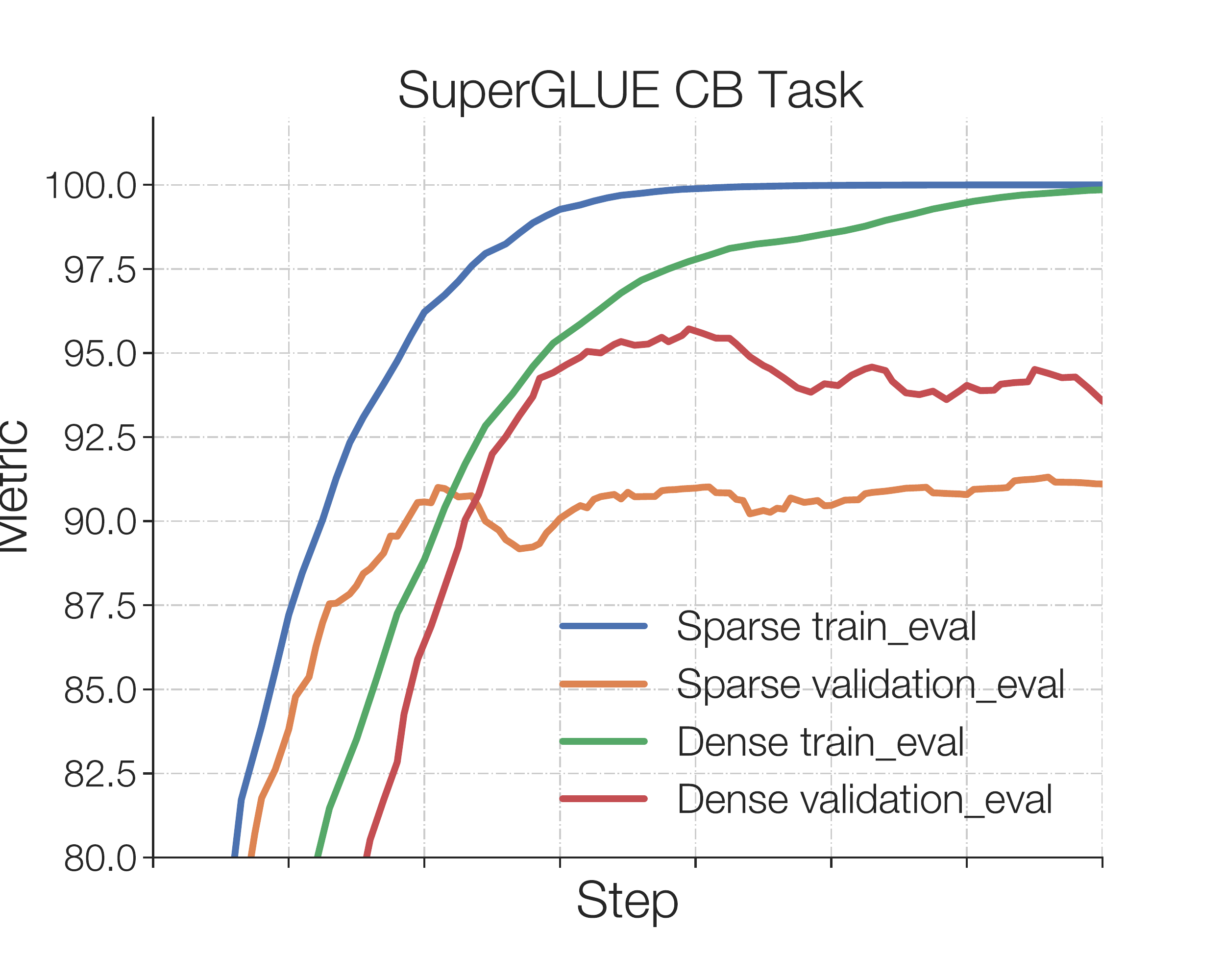}
    \includegraphics[width=0.49\columnwidth]{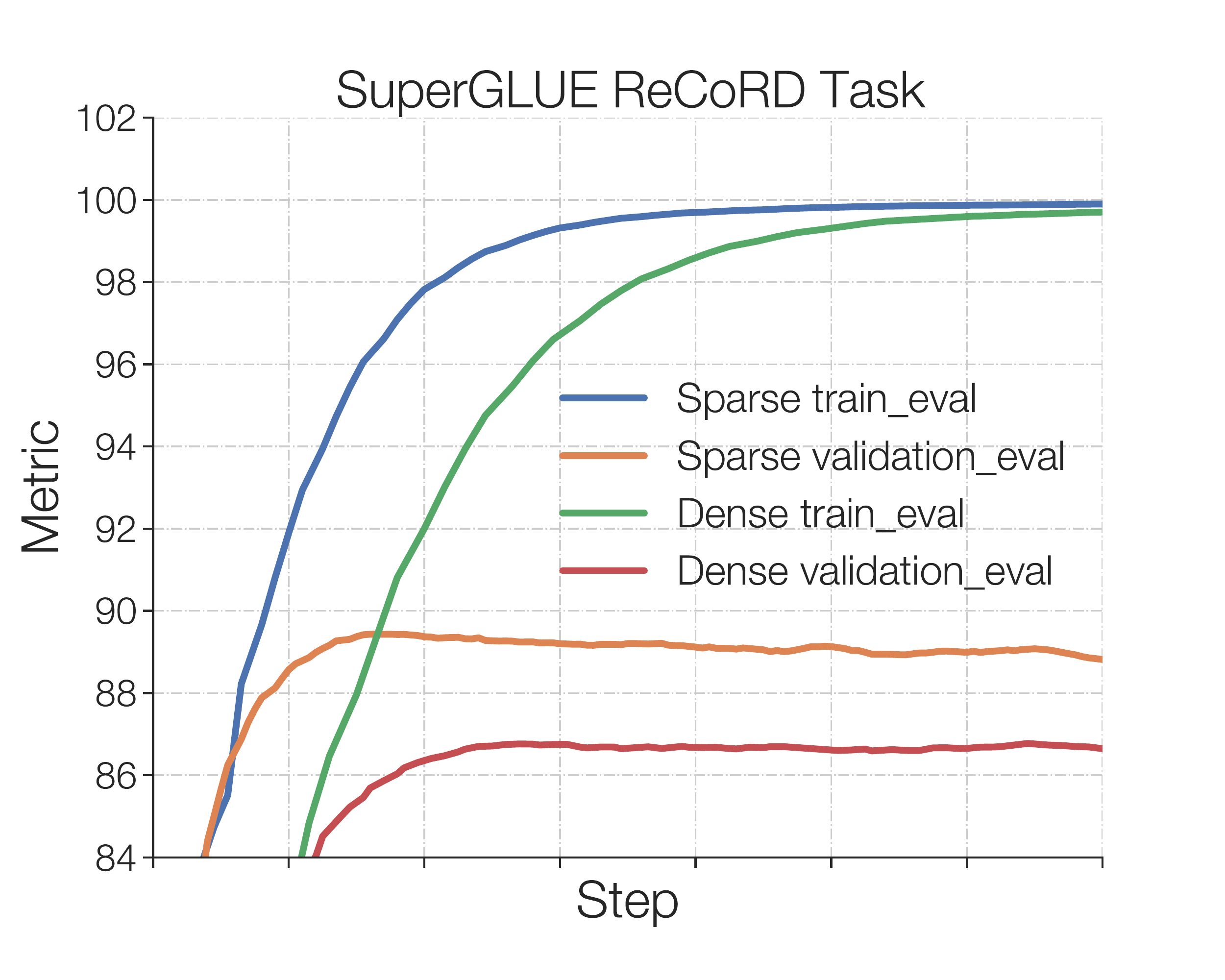}
    \caption{\textbf{Sparse models are prone to overfit.} We plot train and validation curves for our ST-MoE-L and a dense-L models fine-tuned on the CB task (250 train sequences) and ReCoRD (138k train sequences). In both cases, the sparse model learns more quickly on the train partition (blue exceeds green line). However, for the smaller CB task, the dense model outperforms the sparse model on the held-out validation set (red vs.\@ orange). In contrast, on the larger ReCoRD task, the sparse model outperforms the dense model by several percentage points.
    }
    \label{fig:overfitting}
\end{figure}

In Figure \ref{fig:overfitting}, we compare the fine-tuning characteristics of the Dense L and the ST-MoE-L model.
Each model was pre-trained on 500B tokens from the C4 corpus \citep{raffel2019exploring}.
The models are designed to be roughly FLOP matched variants of the T5-Large encoder-decoder models from \cite{raffel2019exploring} with 770M parameters.
The ST-MoE models have 32 experts with an expert layer frequency of $1/4$ (every fourth FFN layer is replaced by an MoE layer). The pre-training and fine-tuning train capacity factor is 1.25 and the eval is 2.0.
We evaluate performance on the held-out validation and train dataset partitions.

\begin{figure}[ht!]
    \centering
    \includegraphics[width=0.45\columnwidth]{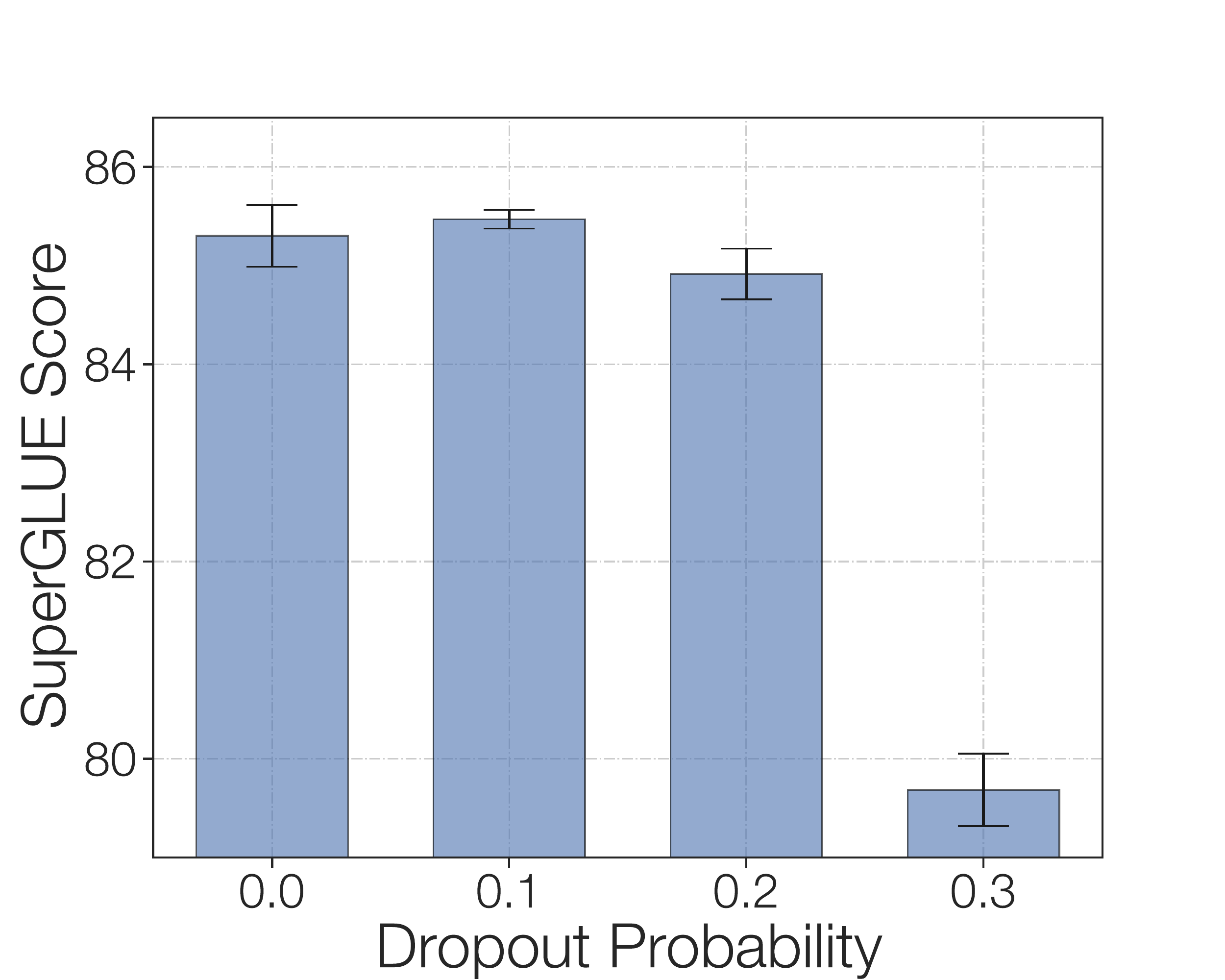}
    \includegraphics[width=0.45\columnwidth]{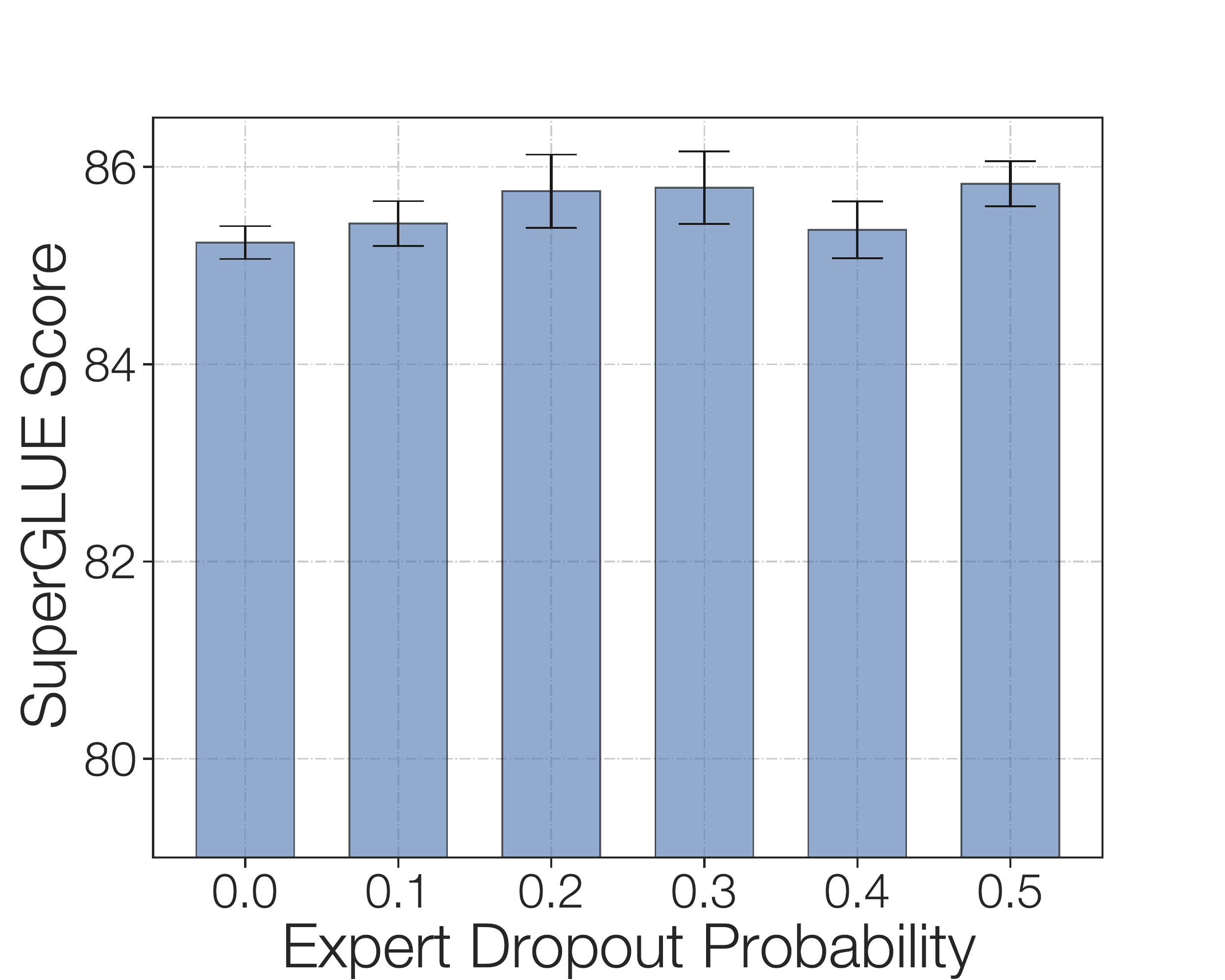}
    \caption{\textbf{Regularization studies of sparse models for fine-tuning}. For each setting, we train three random seeds till convergence on SuperGLUE. We find that increased regularization through dropout provides modest boosts. \textbf{(Left)} demonstrates peak SuperGLUE fine-tuning quality at a global dropout rate of 0.1. Higher values over-regularize and severely hurt quality. \textbf{(Right)} Starting with the best known global dropout rate of 0.1, we selectively increase the expert dropout (an independent dropout rate on the expert hidden activation). This yields further generalization benefits and is in line with the findings of \cite{fedus2021switch}.}
    \label{fig:regularization}
\end{figure}

Across both tasks, the sparse model converges faster to 100\% train set accuracy supporting that sparse models optimize effectively under a data distribution shift.
On the larger task, ReCORD, the validation quality of the sparse model follows the boost in training and significantly exceeds the dense model.
However, on the smaller task, CB, the sparse model lags its dense counterpart on held-out data.
As per the recommendation of \cite{fedus2021switch}, we consider increasing the dropout within the expert hidden state (i.e.\@ expert dropout), but find that at this scale, higher values only moderately improve quality (Figure \ref{fig:regularization}).
We study further improvements to fine-tuning in Section \ref{sec:fine_tune_tricks} and hyperparameter sensitivity in Section \ref{sec:finetuning_sensitivity}.

\subsection{Fine-Tuning a Subset of Model Parameters to Improve Generalization}\label{sec:fine_tune_tricks}

\begin{figure}[ht!]
    \centering
    \includegraphics[width=0.5\columnwidth]{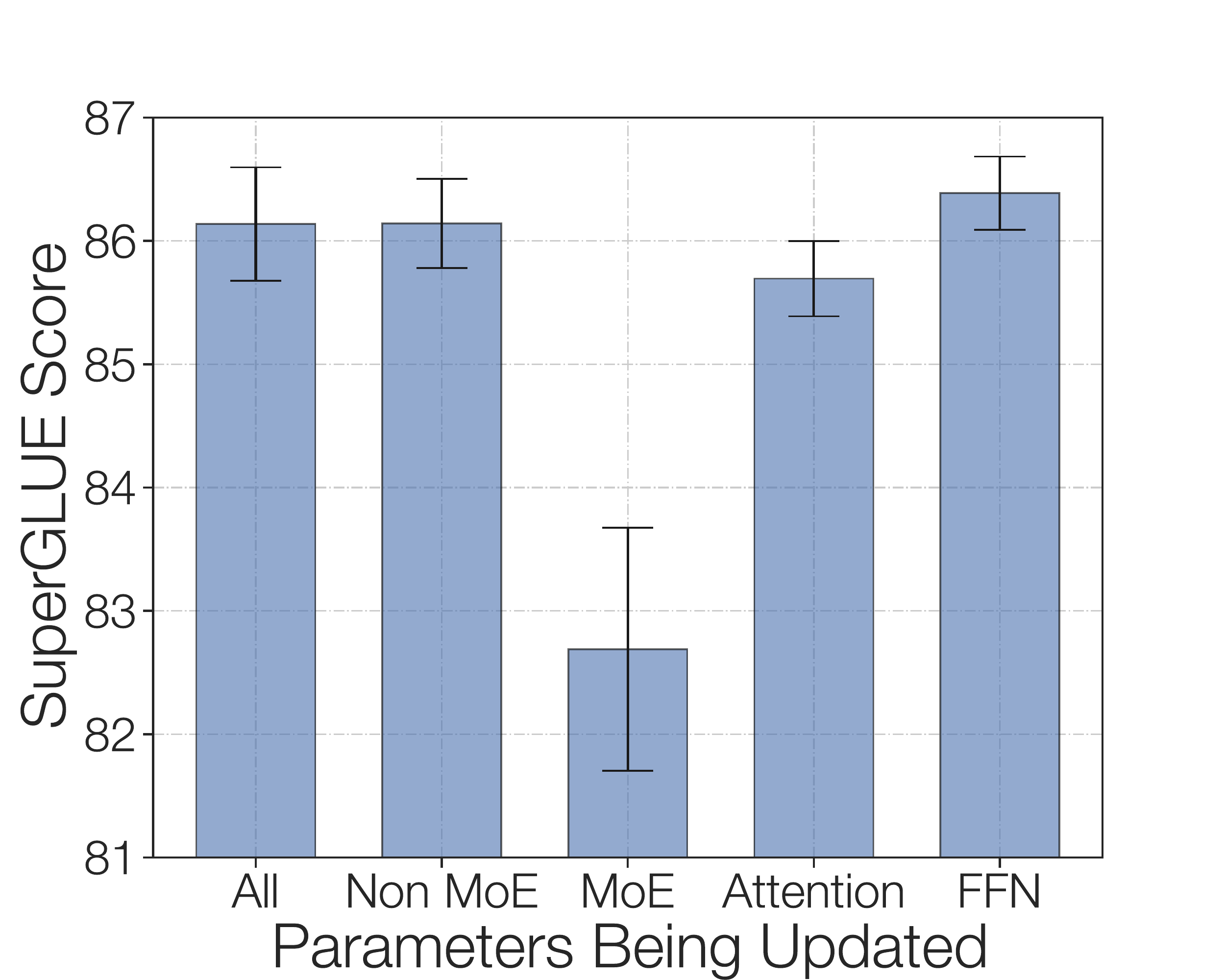}
    \caption{\textbf{Updating only a subset of model parameters during fine-tuning}. To improve the generalization of sparse models and combat overfitting, we fine-tune a subset of the model parameters. All results are with the ST-MoE-L model and are an average of 5 different random seeds. We observe that updating $3/5$ of the subsets of parameters appear to work about the same, while fine-tuning only the MoE parameters results in a drastic quality reduction. }
    \label{fig:hist_subset_ft}
\end{figure}

To combat overfitting we experiment updating only a subset of models parameters during fine-tuning. 
Figure \ref{fig:hist_subset_ft} measures quality for updating 5 different subsets of parameters: all parameters (All), only non MoE parameters (Non MoE), only MoE parameters (MoE), only the self-attention and enc-dec attention parameters (Attention) and only the non MoE FFN parameters (FFN). 

We observe that updating the non MoE  parameters works about as well as updating all the parameters and updating only the FFN parameters works a bit better. Updating only the MoE parameters significantly degrades fine-tuning performance, which is where $\approx$80\% of model parameters are. Only updating the non MoE parameters can be an effective way to speedup and reduce memory for fine-tuning.

We hypothesize that fine-tuning only the MoE parameters  leads to bad performance since expert layers only occur every $1/4$ layers and a token will see at most two experts per layer. Therefore, updating the MoE parameters will affect much fewer layers and FLOPs than updating any other subset of the parameters we tried. Updating only the MoE parameters resulted in a much larger training loss than updating the non MoE parameters, even though there are significantly more  parameters. We further observe that updating all the non-MoE parameters results in a higher training loss than updating all the parameters, but unfortunately this regularization effect didn't translate to better validation performance.

Further, one regularizer we tried was a dropout variant where entire experts were masked out stochastically during training. 
However, this failed to improve generalization in our preliminary studies. 
Appendix~\ref{sec:negative} expands on this experiment and contains other negative results.

\subsection{Sparse and Dense Models Require Different Fine-Tuning Protocols}
\label{sec:finetuning_sensitivity}
\begin{figure}[ht!]
    \centering
    \includegraphics[width=0.45\columnwidth]{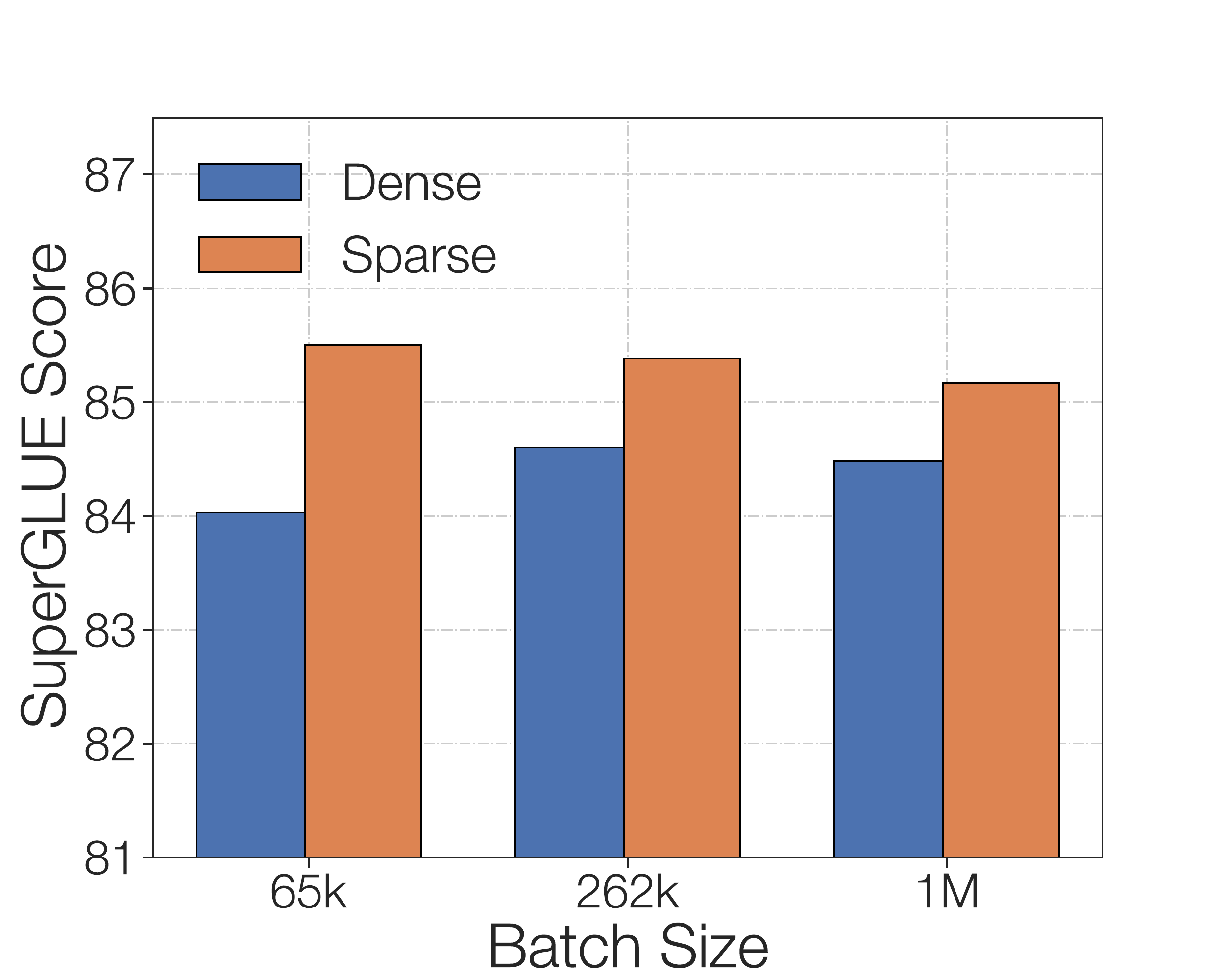}
    \includegraphics[width=0.45\columnwidth]{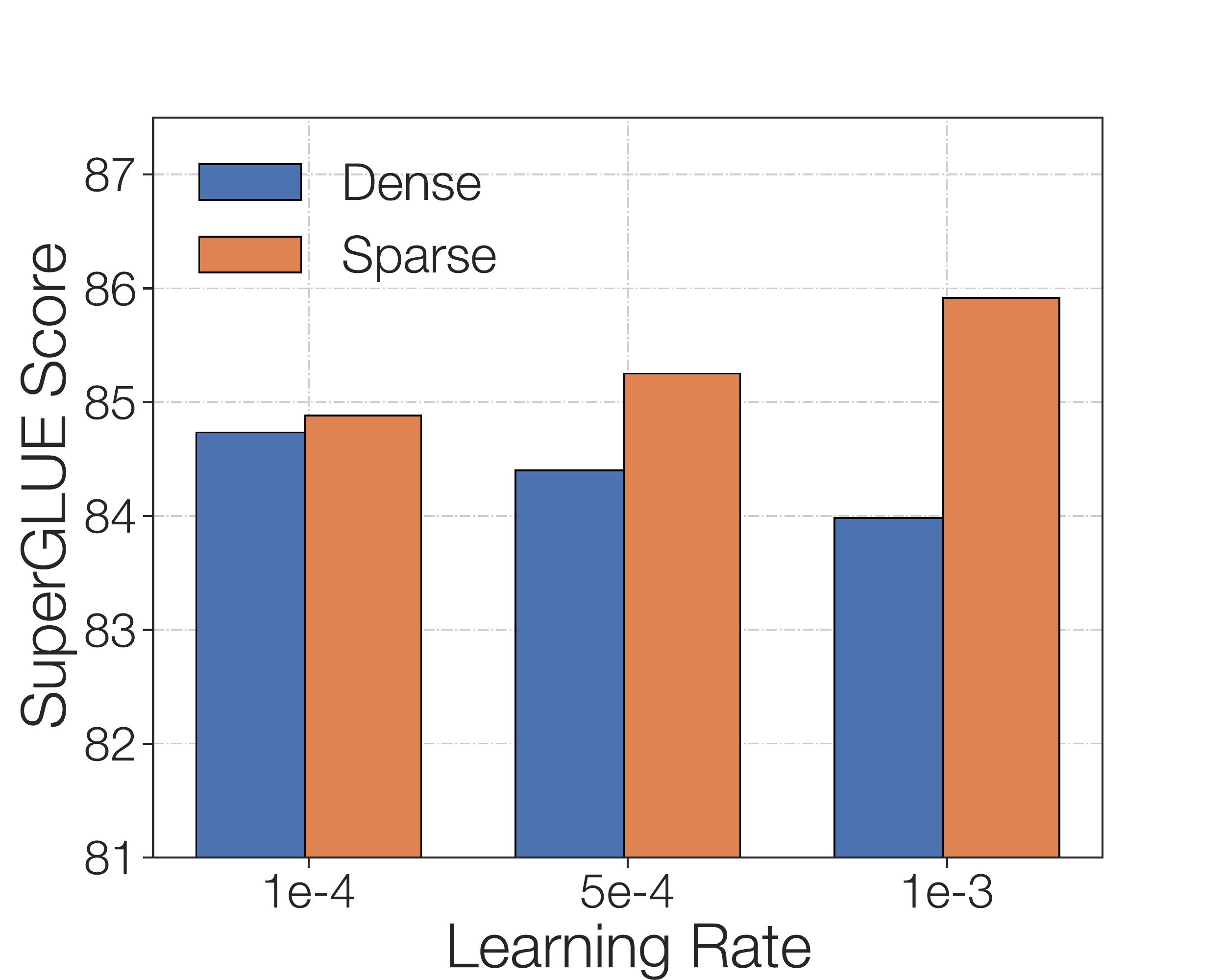}
    \caption{\textbf{Batch size and learning rate sensitivity.} We measure differences and sensitivity to fine-tuning protocols between dense (blue) and sparse (orange) models. Each bar is an average across 6 different runs with different hyperparameters. On SuperGLUE, sparse models benefit from noisier hyperparameters including small batch sizes and high learning rates. Dense models behave nearly oppositely. See Appendix~\ref{appendix: fine_tuning_protocol} for all data. 
    }
    \label{fig:dense_vs_expert_scaling}
\end{figure}

How sensitive are sparse and dense models to the fine-tuning protocol?
We study two hyperparameters: the batch size and the learning rate.
We pretrain a Dense-L and ST-MoE-L on 500B tokens of C4 and then fine-tune on SuperGLUE.
Figure \ref{fig:dense_vs_expert_scaling} summarizes our experiments with the full data presented in Table \ref{tab:fine_tuning_protocol} (Appendix \ref{appendix: fine_tuning_protocol}).
Across all hyperparameter settings, the sparse models (orange) outperform the dense (blue) counterparts -- however, the best setting for each can materially change results.
Sparse and dense models have vastly different performance across different batch sizes and learning rates. 
Sparse models benefit from smaller batch sizes and a higher learning rate.
Consistent with the overfitting hypothesis (Section \ref{sec: overfitting}), both these changes might improve generalization through higher noise in the fine-tuning process. Finally, we point out the importance of correctly tuning the batch size and learning rate during fine-tuning. Simply using the same fine-tuning hyperparameters that worked well for the dense model can mask any pre-training improvements obtained by the sparse model.

\subsection{Sparse Models Are Robust to Dropped Tokens During Fine-Tuning}
Sparse models route tokens to one or more experts at each layer.
To make these models efficient in the SPMD paradigm with modern hardware, the expert capacity (the number of tokens each expert processes) needs to be fixed ahead of time (see Section~\ref{sec:background} for more details). When an expert receives more tokens than its capacity, the extra tokens are dropped --- no computation is applied to those tokens. We again try to prevent this by \textbf{(1)} pre-training with an auxiliary loss that promotes equal amounts of tokens getting sent to each expert and \textbf{(2)} a capacity factor (a hyperparameter) that adds room for extra tokens at each expert. 
We experiment with turning off the auxiliary loss during fine-tuning and using different capacity factors.
Tables \ref{tab:token_dropping_large} reveals a surprising result that fine-tuning quality is not materially impacted by dropping up to 10-15\% of tokens\footnote{Token dropping may be a form of regularization and a more extensive study may be an interesting direction for future work.}.
Studies on ST-MoE-32B corroborate that high capacity factors do not improve fine-tuning quality.
This is in-line with findings of \cite{yang2021m6t} that unequal load balance may not significantly impact model quality.
\begin{table}[H]
    \centering
    \begin{tabular}{c|cccc|cc}
        \toprule
        \textbf{Model} & \textbf{Train CF} & \textbf{Eval CF} & \textbf{Aux Loss} & \textbf{Percent Tokens Dropped} & \textbf{SuperGLUE ($\uparrow$)} \\
        \midrule
        Sparse & 0.75 & 2.0 & Yes & \colorbox{red!50}{10.6\%} & 86.5 $\pm$ 0.21\\
        Sparse & 1.25 & 2.0 & Yes & \colorbox{red!10}{0.3\%} & 86.7\\
        Sparse & 2.0 & 3.0 & Yes & 0.0\% & 85.8\\
        Sparse & 4.0 & 5.0 & Yes & 0.0\% & 86.4 \\
        \midrule
        Sparse & 0.75 & 2.0 & No & \colorbox{red!50}{15.6\%} & 85.7 \\
        Sparse & 1.25 & 2.0 & No & \colorbox{red!30}{2.9\%} & 85.8 \\
        Sparse & 2.0 & 3.0 & No & \colorbox{red!10}{0.4\%} & 85.9 \\
        Sparse & 4.0 & 5.0 & No & 0.0\% & 86.4 \\
        \bottomrule
    \end{tabular}
    \caption{\textbf{Sparse models are robust to dropped tokens when fine-tuning.} We find the fine-tuning quality on SuperGLUE is not impacted significantly across the values explored. Interestingly, dropping 10-15\% of tokens can perform approximately as well as models that drop $<1$\%. We also observe that load balance losses (Aux Loss) improve fine-tuning. The dropped token percentage corresponds to the fraction of dropped tokens across all expert layers at peak validation accuracy.}
    \label{tab:token_dropping_large}
\end{table}

\subsection{Inserting Sentinels Tokens During Fine-Tuning}
Sentinel tokens denote masked sequences in the span-corruption objective \citep{fedus2018maskgan,devlin2018bert}.
This differs from any fine-tuning task we would likely encounter, leading to a domain mismatch between pre-training and fine-tuning.
Table \ref{tab: objectives} illustrates the difference.
We examine whether modifying the fine-tuning task to look more like the pre-training task effects results.

\begin{table}[h!]
    \centering
    \begin{tabular}{lcc}
    \toprule
        \textbf{Objective} & \textbf{Inputs} & \textbf{Targets} \\
        \midrule
        Span Corruption  & I like \textbf{\texttt{<X>}} the pool \textbf{\texttt{<Y>}} day . & \textbf{\texttt{<X>}} going to \textbf{\texttt{<Y>}} on a sunny \\
        Fine-Tuning & What is the capital of Illinois ? & Springfield \\
        Fine-Tuning + Sentinels & What is the capital of Illinois ? \textbf{\texttt{<X>}} & \textbf{\texttt{<X>}} Springfield \\
    \bottomrule
    \end{tabular}
    \caption{\textbf{Inserting sentinels during fine-tuning mimics the pre-training span objective}. We highlight the typical difference between span corruption and fine-tuning. We propose modifying the fine-tuning task to resemble pre-training by inserting sentinel tokens.}
    \label{tab: objectives}
\end{table}

In Table \ref{tab:sentinel} we find that adding sentinel tokens while fine-tuning only improves Grammar Error Correction (GEC) \citep{rothe2021simple}, but not SuperGLUE.
We tried to further reduce the data distribution shift by inserting multiple sentinel tokens (as would be encountered by the model while pre-training), but again found no universal benefit.
However, despite no consistent benefit on held-out data, we find that training convergence is accelerated for both dense and sparse models. 

\begin{table}[h!]
    \centering
    \begin{tabular}{c|c|cc}
        \toprule
        \textbf{Model} & \textbf{Insert Sentinel Tokens} & \textbf{SuperGLUE ($\uparrow$)} & \textbf{GEC ($\uparrow$)} \\
        \midrule
        Dense & & 84.9 $\pm$ 0.33 & 22.3 $\pm$ 0.25 \\
        Dense & \checkmark & 85.1 $\pm$ 0.25 & 22.1 $\pm$ 0.42 \\
        \midrule
        Sparse &  & 86.6 $\pm$ 0.18 & 22.2 $\pm$ 0.04 \\
        Sparse & \checkmark & 86.6 $\pm$ 0.24 & \textbf{22.9} $\pm$ 0.09 \\
        \bottomrule
    \end{tabular}
    \caption{\textbf{Impact of sentinel tokens for fine-tuning}. The addition of sentinel tokens (a similar concept used in \cite{lester2021power}) during fine-tuning has mixed performance on the two tasks we consider. SuperGLUE records the average score and GEC records the exact match. While we find it doesn't improve generalization, sentinel tokens can accelerate training convergence.}
    \label{tab:sentinel}
\end{table}

\section{Designing Sparse Models}
The design of dense models has been guided by the foundational work of \cite{kaplan2020scaling}.
But sparse models pose a myriad of additional questions:
(1) How many experts to use? (2) Which routing algorithm? (3) What value for the capacity factor? (4) How does hardware change these decisions?
In this section, we comment on these and offer recommendations for building Pareto efficient sparse models.
Concurrently, \cite{clark2022unified} provides additional design recommendations including higher layer frequency and top-1 routing as per \cite{fedus2021switch}.

\begin{tcolorbox}[enhanced,attach boxed title to top center={yshift=-3mm,yshifttext=-1mm}, title=\textbf{Designing Sparse Models}, colback=white, colframe=white!75!blue, coltitle=black, colbacktitle=white]
    \begin{enumerate}[leftmargin=*,label=\textbf{\arabic*.}]
        \item In our setup, we recommend top-2 routing with 1.25 capacity factor and at most one expert per core.
        \item The capacity factor can be changed during evaluation to adjust to new memory/compute requirements.
        \item Dense layer stacking and a multiplicative bias can boost quality (Appendix \ref{appendix: architecture}).
    \end{enumerate}
\end{tcolorbox}

\subsection{Setting the Number of Experts}

One of the first questions is the number of experts to use. 
\cite{fedus2021switch} presented the scaling-properties of Switch Transformer which yielded monotonic pre-training benefits (on a step basis) on C4 up to 512-experts, \cite{kim2021scalable} up to 64-experts and \cite{clark2022unified} up to 512-experts.
But the incremental benefit quickly diminishes with many experts ($>$256) or equivalently, with very sparse models ($<$1\%  of experts activated).

However, reflecting on the specific hardware system can further guide this choice.
The compute-to-memory ratio (operational intensity) can serve as an estimate of the efficiency of different operations \citep{williams2009roofline,shazeer2019fast}.
A model is \emph{memory bound} if the time to load tensors to the computing core (e.g.\@ ALU/MMU) greatly exceeds the time required to do the computation on the tensors.
On modern GPUs and TPUs, increasing this compute to memory ratio improves the efficiency.

Returning to sparse expert models, using more than one expert per core increases memory transfer, potentially hurting efficiency.
Increasing the number of experts does not change the computation done (sparse models apply a fixed amount of computation to each input), but increases the memory transfer requirement (additional expert variables must be loaded from device memory).
This \emph{decreases} the compute-to-memory ratio\footnote{As an exercise to the reader, verify the operational intensity of the first expert computation is $\frac{b \cdot h}{b + h \cdot e}$ with $b$ batch size, $h$ hidden dimension, $e$ number of experts.}.
 
On our TPU system, we recommend to one expert (or less) per core.
Our largest models use both data and model parallelism where data parallelism is over ``rows'' and model-parallelism over ``columns'' of the logical mesh.
We use $\leq$ 1 expert per data parallelism row to ensure the compute-to-memory ratio is high and to reduce the cores needed for evaluation and inference.
Furthermore, using less experts lets us allocate more cores to the model parallelism ``column'' to have more FLOPs in our model. 
Appendix \ref{lab:mesh_explanation} explains our mesh layouts for when we have fewer experts than data parallelism rows.

\subsection{Choosing the Capacity Factor and Routing Algorithm}
We generalize top-1 routing \citep{fedus2021switch, roller2021hash} and top-2 \citep{shazeer2017outrageously,lepikhin2020gshard} to study top-$n$ routing where each token is processed by at most $n$ experts.
In this study, all models are pre-trained for 100k steps with 1M tokens per batch and sparse models have 32 experts and are FLOP matched to T5-Large~\cite{raffel2019exploring}. We draw two key conclusions.

First, increasing both the train and eval capacity factors (CF) improves quality as seen by comparing across the segmented blocks of Table \ref{tab:top2_vs_top1}.
For instance, top-1 routing improves by +0.011 neg.\@ log perp.\@ when increasing from 1.0 $\rightarrow$ 1.25 train CF and top-2 routing improves +0.009 increasing from 1.25 $\rightarrow$ 2.0 train CF. 
To provide context for these numbers: tripling the size of a dense model (Dense-L to Dense-XL) yields a +0.090 neg.\@ log perp.\@ boost.
Therefore, these CF boosts are $\sim 1/10^{\text{th}}$ of that magnitude.
But this comes at a cost.
Increasing the capacity factor linearly increases the einsums costs, memory for activations, \texttt{all2all} communication costs, and model-parallelism \texttt{allreduce} communication costs for expert layers\footnote{\texttt{all2all} and \texttt{allreduce} costs depend on the number of devices, batch size, $d_{model}$ and capacity factor, but not on the number of experts.}.

Second, there are small gains of top-$(n+1)$ over top-$n$ routing given a \emph{fixed} capacity factor  (Table~\ref{tab:top2_vs_top1}).
For instance, top-2 routing improves +0.004 over top-1 at train CF of 1.25 or about $1/20^{\text{th}}$ the boost of a dense model tripling.
This revises an earlier recommendation from \cite{fedus2021switch}.
The primary difference between these experimental setups was scale of compute. 
\cite{fedus2021switch} trained 220M-FLOP matched models for 50B tokens.
We find at an 8x larger scale of training (1B-FLOP matched models for 100B tokens) there is instead a small gain to route to more than one expert. 
Furthermore, at the larger experimental scale, the speed difference of top-$n$ versus top-$(n+1)$ routing is negligible.
Speed differences were observed in \cite{fedus2021switch} because the router computation was a larger fraction of the total model computation.

\begin{table}[H]
    \centering
    \small
    \begin{tabular}{c|cc|cc}
         \toprule
         \textbf{Algorithm} & \textbf{Train CF} & \textbf{Eval CF} & \textbf{Neg. Log Perp.} \textbf{($\uparrow$)} \\
         \midrule
         Dense-L & --- & --- & -1.474 \\
         Dense-XL & --- & --- & -1.384 \\
         \midrule
         \rowcolor{gray!15}
         Top-1 & 0.75 & 0.75 & -1.428 \\
         \rowcolor{gray!15}
         Top-1 & 0.75 & 2.0 & -1.404 \\
         Top-2 & 0.75 & 0.75 & -1.424 \\
         Top-2 & 0.75 & 2.0 & -1.402 \\
         \midrule
         \rowcolor{gray!15}
         Top-1 & 1.0 & 1.0 & -1.397 \\
         \rowcolor{gray!15}
         Top-1 & 1.0 & 2.0 & -1.384 \\
         Top-2 & 1.0 & 1.0 & -1.392 \\
         Top-2 & 1.0 & 2.0 & -1.378 \\
         \midrule
         \rowcolor{gray!15}
         Top-1 & 1.25 & 1.25 & -1.378 \\
         \rowcolor{gray!15}
         Top-1 & 1.25 & 2.0 & -1.373 \\
         Top-2 & 1.25 & 1.25 & -1.375 \\
         Top-2 & 1.25 & 2.0 & -1.369 \\
         \midrule
        \rowcolor{gray!15}
         Top-2 & 2.0 & 2.0 & -1.360 \\
         \rowcolor{gray!15}
         Top-2 & 2.0 & 3.0 & -1.359 \\
         Top-3 & 2.0 & 2.0 & -1.360 \\
         Top-3 & 2.0 & 3.0 & -1.356 \\
         \bottomrule
    \end{tabular}
    \caption{\textbf{Comparing capacity factors (CF) and routing algorithms.} Increasing both train and eval CF improves performance. Increasing or decreasing the eval CF gives an additional lever if you have more or less compute at eval time. Next, there are smaller gains of top-$(n+1)$ over top-$n$ routing across capacity factors. Because the quality improves, but the speed slows as the CF increases, the Pareto efficient CF must be determined by the specific hardware system.}
    \label{tab:top2_vs_top1}
\end{table}

The specific hardware-software system will determine the optimal $n$ and capacity factor.
For instance, if the system supports fast \texttt{all2all} and \texttt{allreduce} communications, larger capacity factors and larger $n$ in top-$n$ routing may be optimal.
However, if the \texttt{all2all} and/or \texttt{allreduce} communications are slow, smaller capacity factors may dominate. 
In our case, the hardware-software stack is the TPU and Mesh Tensorflow.
We record the training speed of both our ST-MoE-L and ST-MoE-32B model in Table~\ref{tab: profile_32b} as we increase the train capacity factor.
As the models scale, a higher capacity factor makes the models \emph{increasingly} slower.
The ST-MoE-L does not require model parallelism (it fits within accelerators memory, which implies no additional \texttt{allreduce} communications) making it better suited for high capacity factors than our ST-MoE-32B model.
For our largest model, we therefore continue to use the smaller train capacity factor of 1.25 advocated by \cite{fedus2021switch} for Pareto efficiency, differing from other work which use a larger and more expensive 2.0 capacity factor \citep{lepikhin2020gshard,du2021glam}.

\begin{table}[H]
    \centering
    \small
    \begin{tabular}{l|cc}
         \toprule
         \textbf{Model} & \textbf{Train CF} & \textbf{Step Time (s) ($\downarrow$)} \\
         \midrule
         ST-MoE-L & 1.25 & 2.397 \\
         ST-MoE-L & 2.0 & 2.447 \rlap{($+7\%)$}\\
         \midrule
         ST-MoE-32B & 1.25 & 4.244 \\
         ST-MoE-32B & 2.0 & 4.819 \rlap{($+14\%)$}\\
         \bottomrule
    \end{tabular}
    \caption{\textbf{Profiling sparse models on TPUs.} Increasing the train capacity factor from 1.25 to 2.0 increases the step-time by +7\% for the large (1B) model but by +14\% for our 32B model. As the model size increases, we find the small quality gains of higher train capacity factors from Table \ref{tab:top2_vs_top1} are more than offset by the significant 14\% slow-down. Note: the step time between ST-MoE-L and ST-MoE-32B are not comparable because they used a different number of cores.}
    \label{tab: profile_32b}
\end{table}

Our results in this section focus on top-$n$ routing, but we also experimented with a variety of other routing techniques in Appendix \ref{sec:negative}.
We found most performed similarity or worse compared to top-$n$ routing. 
However we found Batch Prioritized Routing (BPR), introduced in \cite{riquelme2021scaling}, significantly helps performance for capacity factors less than one (Appendix \ref{sec:switch_max}).
We recommend BPR for larger models where \texttt{all2all} and \texttt{allreduce} are more expensive and lower capacity factors are optimal.

\section{Experimental Results}
Given our improvements to training stability, fine-tuning and model design, we start by validating a sparse model approximately FLOP-matched to T5-Large \citep{raffel2019exploring}.
We conclude this section by designing and training a 269B sparse parameter model (FLOP matched to a 32B dense model) which achieves state-of-the-art quality across a wide set of NLP tasks.

We studied the SuperGLUE \citep{wang2019superglue} benchmark throughout this work which consists of tasks including sentiment analysis (SST-2), word sense disambiguation (WIC), sentence similarity (MRPC, STS-B, QQP), natural language inference (MNLI, QNLI, RTE, CB), question answering (MultiRC, RECORD, BoolQ), coreference resolution (WNLI, WSC) and sentence completion (COPA) and sentence acceptability (CoLA).
We often observe good performance on SuperGLUE to correlate with (but not guarantee) performance across many NLP tasks.
We also include a divers set of additional benchmarks.
The CNN-DM \citep{cnn2015moritz} and BBC XSum \citep{narayan2018don} datasets are used to measure the ability to summarize articles.
Question answering is probed with the SQuAD dataset \citep{rajpurkar2016squad} as well as on grade-school science questions in ARC Easy and ARC Reasoning Challenge \citep{clark2018think}.
And as in \cite{roberts2020much}, we evaluate the knowledge of our models by fine-tuning on three closed-book question answer datasets:  Natural Questions \citep{kwiatkowski2019natural}, Web Questions \citep{berant2013semantic} and Trivia QA \citep{joshi2017triviaqa}. 
Closed-book simply refers to questions posed with no supplemental reference or context material. 
To gauge the model's common sense reasoning we evaluate it on the Winogrande Schema Challenge \citep{sakaguchi2020winogrande}.
And finally, we test our model's natural language inference capabilities on the Adversarial NLI Benchmark \citep{nie2019adversarial}.

\subsection{ST-MoE-L}
For simplicity and to cover dozens of tasks easily, we train on \emph{mixtures} of the tasks listed rather than separately fine-tuning a model on each task.
However, because the tasks vary in size considerably, equally sampling per the number of examples would over-sample large tasks and under-sample small ones.
We therefore mix each task in proportion to the number of examples in its `train' split (up to some \texttt{max\_num\_examples=65536}) as in \cite{raffel2019exploring}.
This means that tasks containing more than 65536 training examples are weighted as if they only contain \texttt{max\_num\_examples}.

\begin{table*}[ht!]
    \centering
    \begin{tabular}{llcccc}
        \toprule
        \textbf{Name} & \textbf{Metric} & \textbf{Split} & \textbf{Dense-L ($\uparrow$)} & \textbf{ST-MoE-L ($\uparrow$)} & \textbf{Gain (\%)} \\
        \midrule
        SQuADv2 & F1 & dev & 94.0 & \textbf{94.5}  & $+1\%$ \\
        SQuADv2 & acc & dev & 87.6 & \textbf{88.1}  & $+1\%$ \\
        \addlinespace
        SuperGLUE & avg & dev & 85.1 & \textbf{87.4} & $+3\%$ \\
        \hspace{2mm}BoolQ & acc & dev & 87.1 & \textbf{88.6} & $+2\%$ \\
        \hspace{2mm}Copa & acc & dev & 83.0 & \textbf{91.0} & $+10\%$ \\
        \hspace{2mm}RTE & acc & dev & 91.0 & \textbf{92.1} & $+1\%$ \\
        \hspace{2mm}WiC & acc & dev & 70.4 & \textbf{74.0} & $+5\%$ \\
        \hspace{2mm}MultiRC & F1 & dev & 83.9 & \textbf{86.0} & $+3\%$ \\
        \hspace{2mm}WSC & acc & dev & \textbf{95.2} & 93.3 & $-2\%$ \\
        \hspace{2mm}ReCoRD & acc & dev & 85.7 & \textbf{88.9} & $+4\%$ \\
        \hspace{2mm}CB & acc & dev & \textbf{100} & 98.2 & $-2\%$ \\
        \addlinespace
        XSum & ROUGE-2 & dev & 19.9 & \textbf{21.8} & $+10\%$ \\
        CNN-DM & ROUGE-2 & dev & 20.3 & \textbf{20.7} & $+2\%$ \\
        \addlinespace
        WinoGrande (XL) & acc & dev & 75.4 & \textbf{81.7} & $+8\%$ \\
        ANLI (R3) & acc & dev & 54.3 & \textbf{57.3} & $+6\%$ \\
        ARC-Easy & acc & dev & 63.5 & \textbf{75.4} & $+19\%$ \\
        ARC-Challenge & acc & dev & 50.2 & \textbf{56.9} & $+13\%$ \\
        \addlinespace
        Closed Book TriviaQA & acc & dev & 28.1 & \textbf{33.8} & $+20\%$ \\
        Closed Book NatQA & acc & dev & 27.2 & \textbf{29.5} & $+8\%$ \\
        Closed Book WebQA & acc & dev & 30.5 & \textbf{33.2} & $+9\%$ \\
        \bottomrule
    \end{tabular}
        \caption{\textbf{Fine-tuning performance of FLOP-matched dense and sparse models.} Comparison of the dense-L baseline and the sparse FLOP-matched version (higher numbers better). We observe consistent gains across diverse tasks, using approximately the same amount of computation. The only two tasks without improvement from the sparse model are the two smallest: CB with 250 training examples and WSC with 259.}
        \label{tab: multitask}
\end{table*}

Table \ref{tab: multitask} summarizes the quality of a dense T5-Large (L) model and sparse model with approximately the same number of FLOPs pre-trained for 500k steps with a 1M batch size (524B tokens) on the C4 dataset \citep{raffel2019exploring}. The sequence length for the encoder was 512 and 114 for the decoder.
We observe improvements on the validation (dev) sets across a wide array of tasks examining natural language understanding, question answering, and summarization.
As seen in \cite{fedus2021switch}, striking gains are observed in closed book question answering \citep{roberts2020much}. 

Also, in support of the overfitting hypothesis presented in Section \ref{sec: overfitting}, we observe two of the smallest tasks CB and WSC (250 and 259 training examples, respectively), are the only ones where the sparse model does not yield gains over its dense counterpart.
This again suggests that improved forms of regularization for sparse models may unleash greater performance.

\subsection{ST-MoE-32B}
With quality validated at the scale of T5-Large, we seek to push the capabilities of sparse models through the ST-MoE-32B.
When designing this, we sought a balance between FLOPs and parameters.
High-FLOP sparse models were previously unstable in \cite{fedus2021switch} in our setting (i.e.\@ encoder-decoder models, Adafactor optimizer), but the router z-loss enabled us to proceed.
For computational efficiency, we expanded the hidden size of the experts ($d_{ff}$ in Table \ref{tab: model_params} below)\footnote{\texttt{allreduce} activation communications introduced through model parallelism are independent of the hidden size, but not the model dimension, making it a good choice to increase.}.
Finally, we increased the $d_{kv}$ to 128 for better performance on our hardware. 
The most salient changes are fewer overall parameters and more FLOPs per token relative to both Switch-C and Switch-XXL. 
Our ST-MoE-32B has ``only'' 269B parameters and is approximately FLOP-matched to a dense Transformer with 32B parameters. 
The reduced parameter count from Switch-C and Switch-XXL eases the burden for both serving and fine-tuning.
Finally, we use the sparse-dense stacking described in Appendix \ref{appendix: architecture}.

\begin{table}[ht!]
\begin{center}
\resizebox{\columnwidth}{!}{
    \begin{tabular}{c|cccccc}
        \toprule
        \textbf{Model} & \textbf{Parameters} & \textbf{FLOPs/seq} & \textbf{$d_{model}$} & \textbf{$\text{FFN}_{\text{GEGLU}}$} & \textbf{$d_{ff}$} &  \textbf{$d_{kv}$}  \\ \midrule
        Dense-L & 0.8B & 645B & 1024  & \checkmark & 2816 & 64\\
        T5-XXL     & 11.1B  & 6.3T & 4096  & \checkmark & 10240 & 64  \\
        \midrule
        Switch-XXL & 395B  & 6.3T & 4096 & \checkmark & 10240 & 64  \\
        Switch-C   & 1571B & 890B & 2080  &  & 6144 & 64 \\
        \midrule
        ST-MoE-L  & 4.1B & 645B & 1024  & \checkmark & 2816 & 64 \\
        ST-MoE-32B & 269B & 20.2T & 5120 & \checkmark & 20480 & 128 \\
        \bottomrule
        \\
        \toprule 
        \textbf{Model} & \textbf{Num. Heads} & \textbf{Num. Layers} & \textbf{Num. Experts} & \textbf{Expert Layer Freq.} & \textbf{Sparse-Dense} & \\ \midrule
        Dense-L & 16  & 27 & -- & -- &  &  \\
        T5-XXL     & 64 & 24 & --   & -- &   & \\
        \midrule
        Switch-XXL & 64 & 24 & 64   & $1 / 4$ &  &  \\
        Switch-C   & 32  & 15 & 2048 & $1 / 1$ &  &  \\
        \midrule
        ST-MoE-L   & 16  & 27 & 32 & $1 / 4$ & \checkmark &  \\
        ST-MoE-32B & 64 & 27 & 64 & $1 / 4$ & \checkmark & \\
        \bottomrule
    \end{tabular}
}
\caption{\textbf{Model comparisons.} A comparison of the Dense-L and T5-XXL, the two largest Switch Transformer variants (Switch-XXL and Switch-C), and the ST-MoE-L and ST-MoE-32B. $d_{model}$ refers to the model hiddenstate size and $d_{ff}$ is the internal size of the FFN layer. $d_{kv}$ is the dimension of each attention head. Expert Layer Freq.\@ is the fraction of FFN layers replaced with a sparse layer. Sparse-Dense refers to the architectural variant described in Appendix \ref{appendix: architecture}.}
\label{tab: model_params}
\end{center}
\end{table}

We pre-train for 1.5T tokens on a mixture of English-only C4 dataset \citep{raffel2019exploring} and the dataset from GLaM \citep{du2021glam} summarized in Appendix \ref{app: pretrain_data}.
We use 1M tokens per batch, the Adafactor optimizer with default hyperparameters, and a learning rate warm-up of 10k steps followed by inverse square root decay.
Our model follows the initialization scheme proposed in \cite{fedus2021switch}.

\begin{table*}[!ht]
    \centering
    \begin{tabular}{llc|ccc|c}
        \toprule
        & & & \multicolumn{3}{c}{\bf Previous Best ($\uparrow$)} & {\bf Ours ($\uparrow$)}\\
        \cmidrule(lr){4-6} \cmidrule(lr){7-7} 
        \textbf{Name} & \textbf{Metric} & \textbf{Split} & Zero-Shot & One-Shot & Fine-Tune  & Fine-Tune \\
        \midrule
        SQuADv2 & F1 & dev & 68.3$^e$ & 70.0$^e$ & $96.2^a$ & \textbf{96.3} \\
        SQuADv2 & acc & dev & 62.1$^e$ & 64.6$^e$ & \textbf{91.3}$^a$ & 90.8 \\
        \addlinespace
        SuperGLUE & avg & test & -- & -- & 90.9 & \textbf{91.2} \\
        \hspace{2mm}BoolQ & acc & dev/test & 83.0$^e$ & 82.8$^e$ & 92.0 & \textbf{92.4} \\
        \hspace{2mm}Copa & acc & dev/test & 91.0$^d$ & 92.0$^e$ & 98.2 & \textbf{99.2} \\
        \hspace{2mm}RTE & acc & dev/test & 68.8$^e$ & 71.5$^e$ & \textbf{94.1} & 93.5 \\
        \hspace{2mm}WiC & acc & dev/test & 50.5$^e$ & 52.7$^e$ & \textbf{77.9} & 77.7 \\
        \hspace{2mm}MultiRC & F1 & dev/test & 72.9$^d$ & 72.9$^d$ & 88.6 & \textbf{89.6} \\
        \hspace{2mm}WSC & acc & dev/test & 84.9$^e$ & 83.9$^e$ & \textbf{97.3} & 96.6 \\
        \hspace{2mm}ReCoRD & acc & dev/test & 90.3$^e$ & 90.8$^e$ & \textbf{96.4} & 95.1 \\
        \hspace{2mm}CB & acc & dev/test & 46.4$^d$ & 73.2$^e$ & \textbf{99.2} & 98.0 \\
        \addlinespace
        XSum & ROUGE-2 & test & -- & -- & 24.6$^h$ & \textbf{27.1} \\
        CNN-DM & ROUGE-2 & test & -- & -- & 21.6$^a$ & \textbf{21.7} \\
        \addlinespace
        WinoGrande XL & acc & dev & 73.4$^e$ & 73.2$^d$ & -- & 96.1 \\
        ANLI R3 & acc & test & 40.9$^e$ & 40.8$^e$ & 53.4 & \textbf{74.7}\\
        \addlinespace
        ARC-Easy & acc & test & 71.9$^e$ & 76.6$^e$ & 92.7$^g$ & \textbf{95.2} \\
        ARC-Challenge & acc & test & 51.4 & 53.2 & 81.4$^g$ & \textbf{86.5} \\
        \addlinespace
        CB TriviaQA & em & dev & 68.0$^e$ & \textbf{74.8}$^e$ & 61.6$^b$ & 62.3 \\
        CB NatQA & em & test & 21.5$^e$ & 23.9$^e$ & $41.5^c$ &  \textbf{41.9} \\
        CB WebQA & em & test & 38.0$^f$ & 25.3 & $42.8^b$ & \textbf{47.4}\\
        \bottomrule
    \end{tabular}
        \caption{\textbf{ST-MoE-32B versus previous best for inference-only techniques and fine-tuned models.} A split of ``dev/test'' refers to dev split for Zero-Shot and One-Shot and test split for Fine-Tune quality. Data not available filled in with ``--''. Superscript letters denote the result: $^a$: \cite{raffel2019exploring} $^b$: \cite{roberts2020much} $^c$: \cite{karpukhin2020dense}, $^d$: \cite{brown2020language}, $^e$: \cite{du2021glam}, $^f$: \cite{wang2021ernie}, $^g$: UnifiedQA + ARC MC/DA + IR, $^h$: \cite{zhang2020pegasus}.
        }
        \label{tab:main-results}
\end{table*}

Table \ref{tab:main-results} evaluates our ST-MoE-32B model against previous state-of-the-art approaches using inference-only (zero-shot, one-shot) as well as fine-tuning.
On SuperGLUE, our model improves upon the prior state-of-the-art model, achieving an average score of 91.2 on the test server (93.2 validation accuracy) which is over one percentage point beyond estimated human capability.
For both summarization datasets, XSum and CNN-DM, our model achieves state-of-the-art without additional changes to training or fine-tuning \citep{raffel2019exploring,liang2021rdrop}.
ST-MoE-32B improves the current state-of-the-art on the test server submissions for both ARC Easy (92.7 $\rightarrow$ 94.8) and ARC Challenge (81.4 $\rightarrow$ 86.5).
On two of the three closed book QA tasks, we improve over the prior state-of-the-art.
Closed book WebQA achieves a 47.4 accuracy (prior best of 42.8 from \cite{roberts2020much} and exceeds results from the zero-shot performance of the ERNIE 3.0 Titan 260B dense parameter model \citep{wang2021ernie}).
Closed book NatQA improves to 41.9 accuracy (prior best of 41.5 from \cite{karpukhin2020dense}). 
We find significant improvements on adversarially constructed datasets (ANLI R3 and WinoGrande XL).
ANLI R3 \citep{nie2019adversarial} improves the state-of-the-art to 74.7 (prior best of 53.4).  

We note some weaknesses in our model.
ST-MoE-32B has lackluster performance on the small SQuAD dataset, with an exact match score of 90.8 which falls short of the older benchmark set by the T5-XXL of 91.3.
Furthermore, while setting a new state-of-the-art for SuperGLUE in aggregate, certain tasks, including small ones like CB, WSC, fail to improve.
Finally, on closed book Trivia QA, our model improves over the fine-tuned baseline with SSM from \cite{roberts2020much}, but fails to produce gains over both GPT-3 and GLAM.

While not the focus of this paper, we present the quality differential between recent advances in inference-only techniques like few-shot learning and fine-tuning on these tasks (GPT-3 \citep{brown2020language}, GLAM \citep{du2021glam} and Gopher \citep{rae2021scaling}). 
As expected and observed previously, fine-tuning outperforms zero/one-shot learning, but has the disadvantage of requiring additional training and different models for each task.

\section{Tracing Tokens Through the Model}
Thus far we have presented quantitative measures and performance metrics.
We change tack to explore \emph{qualitative} features by visualizing how tokens are routed among the experts.
We do so by passing a batch of tokens to the model and manually inspecting token assignment at each layer.
We consider our ST-MoE-L model pre-trained either on the monolingual C4 corpus \citep{raffel2019exploring} or on the multilingual mC4 corpus \citep{xue2020mt5}.
On both the encoder and the decoder, the model has six sparse layers, each with 32 experts.

\begin{tcolorbox}[enhanced,attach boxed title to top center={yshift=-3mm,yshifttext=-1mm}, title=\textbf{Preliminaries}, colback=white, colframe=white!75!blue, coltitle=black, colbacktitle=white]
The span corruption objective is to recover spans of variable-length contiguous segments masked out in the inputs. 
This is formatted as: \\

\textit{\textbf{Inputs:} I went to \textless extra\_id\_0\textgreater \: to buy \textless extra\_id\_1\textgreater \:} \\

\textit{\textbf{Targets:} \textless extra\_id\_0\textgreater \: the store \textless extra\_id\_1\textgreater \: milk} \\

In our encoder-decoder architecture, the inputs will be passed to the encoder and targets to the decoder. 
\end{tcolorbox}

Each group of tokens is routed jointly with load balancing across experts incentivized by an auxiliary loss as proposed in \cite{shazeer2017outrageously} (see Appendix \ref{appendix:load_balance} for details).
Tokens compete for expert assignment against other tokens in their group, rather than the entire batch, and expert specialization is heavily influenced by the distribution of tokens in each group. The notion of groups is introduced to limit the cost of dispatching and gathering the correct tokens to the correct experts.

\subsection{Encoder Experts Exhibit Specialization}
Our first observation is that, at each layer, at least one expert specializes in sentinel tokens (mask tokens that represent blanks to fill-in).
Additionally, some encoder experts exhibit clear specialization, with some experts primarily operating on punctuation, verbs, proper names, counting, etc.
Table~\ref{tab:encoder_expert_specialization} presents a few notable example of specialization across encoder experts.
And while we find many instances of specialization, these have been specifically extracted from many examples without a clear semantic or syntactic specialization.  

\begin{table}[h!]
\small
\begin{center}
\renewcommand{\arraystretch}{1.2}
\begin{tabular}{l|l|l}
    \toprule
    \textbf{Expert specialization} & \textbf{Expert position} & \textbf{Routed tokens} \\
    \midrule
    \textbf{Sentinel tokens} & Layer 1 & been \textless extra\_id\_4\textgreater \textless extra\_id\_7\textgreater floral to \\
    & & \textless extra\_id\_10\textgreater \textless extra\_id\_12\textgreater \textless extra\_id\_15\textgreater \\
    & & \textless extra\_id\_17\textgreater \textless extra\_id\_18\textgreater \textless extra\_id\_19\textgreater ... \\
    & Layer 4 & \textless extra\_id\_0\textgreater \textless extra\_id\_1\textgreater \textless extra\_id\_2\textgreater \\
    & & \textless extra\_id\_4\textgreater \textless extra\_id\_6\textgreater \textless extra\_id\_7\textgreater \\
    & & \textless extra\_id\_12\textgreater \textless extra\_id\_13\textgreater \textless extra\_id\_14\textgreater ... \\
    & Layer 6 & \textless extra\_id\_0\textgreater \textless extra\_id\_4\textgreater \textless extra\_id\_5\textgreater \\
    & & \textless extra\_id\_6\textgreater \textless extra\_id\_7\textgreater \textless extra\_id\_14\textgreater \\
    & & \textless extra\_id\_16\textgreater \textless extra\_id\_17\textgreater \textless extra\_id\_18\textgreater ... \\
    \midrule
    \textbf{Punctuation} & Layer 2 & , , , , , , , , , - , , , , , ). )  \\
    & Layer 6 &  , , , , , : . : , \& , \& \& ? \& - , , ? , , , . \textless extra\_id\_27\textgreater \\
    \midrule
    \textbf{Conjunctions and articles} & Layer 3 & The the the the the the the the the The the the \\
    & & the the the The the the the \\
    & Layer 6 & a and and and and and and and or and a and . \\
    & & the the if ? a designed does been is not \\
    \midrule
    \textbf{Verbs} & Layer 1 & died falling identified fell closed left posted lost felt \\
    & & left said read miss place struggling falling signed died \\
    & & falling designed based disagree submitted develop \\
    \midrule
    \textbf{Visual descriptions} & Layer 0 & her over her know dark upper dark outer \\
    \textit{color, spatial position} & & center upper blue inner yellow raw mama \\
    & & bright bright over open your dark blue \\
    \midrule
    \textbf{Proper names} & Layer 1 & A Mart Gr Mart Kent Med Cor Tri Ca Mart \\
    & & R Mart Lorraine Colin Ken Sam Ken Gr Angel A \\
    & & Dou Now Ga GT Q Ga C Ko C Ko Ga G \\
    \midrule
    \textbf{Counting and numbers} & Layer 1 & after 37 19. 6. 27 I I Seven 25 4, 54 I two dead we \\
    \textit{written and numerical forms} & & Some 2012 who we few lower each \\
    \bottomrule
\end{tabular}
\vspace{-0.2cm}
\caption{
\textbf{Notable examples of specialization in encoder experts}.
We find experts that specialize in punctuation, conjunctions \& articles, verbs, visual descriptions, proper names, counting \& numbers.
Across all layers (not shown), we observe experts that primarily operate on sentinel tokens (marked as \textless extra\_id\_x\textgreater).
Note that a SentencePiece model \citep{kudo2018sentencepiece} will split a token if it doesn't exist in the vocabulary, e.g.\@ Kenneth may become Ken, ne, th.}
\label{tab:encoder_expert_specialization}
\end{center}
\vspace{-0.15cm}
\end{table}

\subsection{Decoder Experts Lack Specialization}
In contrast, \emph{expert specialization is far less noticeable in the decoder}.
Not only are sentinel tokens routed somewhat uniformly across decoder experts  (see Table~\ref{tab:sentinel_entropy}), but we also do not observe meaningful specialization (semantics or syntax) in decoder experts.

We hypothesize that this lack of meaningful expert specialization is caused by the \emph{distribution of target tokens induced by the span corruption objective}.
In particular, \textbf{(a)} a smaller number of tokens are routed jointly in the decoder due to longer sequence lengths in the encoder (e.g.\@ group size is 2048 in the encoder vs 456 in the decoder in our setup) and \textbf{(b)} a higher proportion of tokens are sentinel tokens in the decoder.
As a result, target tokens in each group typically cover a smaller semantic space (compared to the encoder), perhaps explaining the lack of expert specialization in the decoder.
This intricate interplay between the architecture and the training objective invites further research on better leveraging sparsity and expert specialization in the decoder. Alternatively, future work could study simply removing the experts in the decoder layer, which also confers benefits during autoregressive decoding~\citep{kudugunta2020exploring}.

\begin{table}[h!]
    \centering
    \begin{tabular}{l|cccccc|c}
        \toprule
        & Layer 1 & Layer 2 & Layer 3 & Layer 4 & Layer 5 & Layer 6 & Uniform (32-experts) \\
        \midrule
        Encoder & 2.2 & 1.8 & 1.6 & 1.7 & 1.7 & 1.2 & 3.5\\
        Decoder & 3.4 & 3.4 & 3.4 & 3.4 & 3.4 & 3.4 & 3.5 \\
        \bottomrule
    \end{tabular}
    \vspace{-0.3cm}
    \caption{\textbf{Entropy of routed sentinel tokens across encoder and decoder layers.} 
    We support our qualitative observation that encoder experts specialize, but decoder expert don't by computing the entropy over the routing for sentinel tokens. The encoder routing entropy is low, but the decoder router is high entropy, and nearly equal to uniform routing. Because each layer has 32-experts, a completely uniform distribution has entropy of 3.5. 
    }
    \label{tab:sentinel_entropy}
\end{table}

\subsection{Multilingual Experts Specialize, But Not by Language}
We next consider a \emph{multilingual} sparse model pretrained on a mixture of different languages and inspect the expert specialization in the encoder.
As in the monolingual case, we find strong evidence of expert specialization.
Table~\ref{tab:multilingual_expert_specialization} presents some examples of experts specializing in sentinel tokens, numbers, conjunctions \& articles and proper names.

\begin{table}[h!]
\small
\begin{center}
\renewcommand{\arraystretch}{1.2}
\begin{CJK*}{UTF8}{gbsn}
\begin{tabular}{l|l}
    \toprule
    \textbf{Expert specialization} & \textbf{Routed tokens} \\
    \midrule
    \textbf{Sentinel tokens} & to \textless extra\_id\_6\textgreater to til \textless extra\_id\_9\textgreater \\
    & \textless extra\_id\_10\textgreater to \textless extra\_id\_14\textgreater \textless extra\_id\_17\textgreater\\ 
    & \textless extra\_id\_19\textgreater  \textless extra\_id\_20\textgreater  \textless extra\_id\_21\textgreater ... \\
    \midrule
    \textbf{Numbers} & \$50 comment .10.2016 ! 20 20 3 ! 5 1. ! 91 ? né ? \\
    & 2 17 4 17 11 17 8 \& 11 \& 22:30 02 2016. ) iOS \\
    \midrule
    \textbf{Conjunctions \& Articles } & of of of their their of any this this your your am von \\
    & this of Do of of This these our 的 的 于 的 在 的 在 的 \\
    & le les Le la di la sur sur 136 sur の の する の という の し \\
    \midrule
    \textbf{Prepositions \& Conjunctions} & For for or for for or for from because https during https \\
    & 并 与 和 par c Pour à a par trè pour pour pour pour pour c と や のに \\
    & で で で なので - and and + c between and and \\
    \midrule
    \textbf{Proper names} & Life Apple  iOS  A IGT 众 莫 HB \\
    & F HB A K A OPP OK HB A Gia C Gia C P Scand Wi \\
    & G H Z PC G Z ハイ PC G Ti CPU PC PC A キット OS \\
    \bottomrule
\end{tabular}
\end{CJK*}
\vspace{-0.2cm}
\caption{
\textbf{Examples of specialization in multilingual experts (encoder)}.
Multilingual experts also exhibit specialization, which sometimes spans across different languages (e.g.\@ "for" and "pour").
Experts trained on multilingual mixtures do not exhibit language specialization.
}
\label{tab:multilingual_expert_specialization}
\end{center}
\vspace{-0.15cm}
\end{table}

One might expect experts to specialize in languages, which appears as a natural criterion for divvying up batches of data among experts.
However, we find no evidence of language specialization (see Table~\ref{tab:multilingual_expert_specialization}).
Routers instead pass tokens from English, Japanese, French and Chinese indiscriminately and the experts appear to be \emph{multilingual}.
But this lack of language specialization is less surprising when considering the mechanism of token routing and load balancing.
Since each group of tokens may only contain one, to at most a few, languages (a group usually consists of 2-4 sequences in our setup), then all experts are encouraged to handle tokens from all languages.
We experimented with a global load balance loss, however, this usually results in worse load-balance and worse model performance, so we leave further improving multilingual expert models as an area of open work (Section \ref{sec:discussion}).

Our visualization reveals apparent specialization learned in our models (Tables \ref{tab:encoder_expert_specialization}, \ref{tab:multilingual_expert_specialization}) for the encoder layers.
Other expert specializations were also observed in the appendix of \cite{shazeer2017outrageously}. 
However, this leads to an interesting question of how architectures that eliminate learned routing \cite{roller2021hash,zuo2021taming} appear to perform well.
An extensive study of the scaling properties of learned versus random routing could prove helpful as future work and help guide us to a better understanding of routing behavior.

\section{Related Work}
Mixture-of-Experts (MoE) date back at least three decade history to the work of \cite{jacobs1991adaptive,jordan1994hierarchical}.
In initial concepts, the MoE defined the entire neural network akin to ensemble methods.
But later \cite{eigen2013learning} extended the idea of including MoE as a \emph{component} as part of deeper networks.
\cite{shazeer2017outrageously} then scaled this idea to a 137B parameter model to achieve state-of-the-art in machine translation.
Most of the later work (including ours) follows this MoE as a component approach.

\textbf{Scale in natural language processing.}
The remarkable success of scale in natural language processing \citep{kaplan2020scaling, brown2020language} has reinvigorated MoE research evidenced by a surge of recent work \citep{lepikhin2020gshard, fedus2021switch, yang2021m6t, kim2021scalable, du2021glam, artetxe2021efficient, zuo2021taming, clark2022unified}.
Sparse expert models have been proposed as a method to achieve the results of large-scale dense models, more efficiently.
\cite{fedus2021switch} showed a 4x pre-train speed-up over T5-XXL \citep{raffel2019exploring} and \cite{du2021glam} matched the quality of GPT-3 \citep{brown2020language} using only $1/3$ of the energy.
And in the span of the last twelve months, a milestone of efficiently training trillion parameter deep neural networks has been achieved by multiple groups
\citep{fedus2021switch,yang2021m6t,du2021glam}, and most recently, \cite{lin2021m610t} introduced techniques to train a 10T parameter model.
One side note is that the recent significant successes of sparse expert models have often been in settings with a lot of data and no distribution shift -- two examples being language modeling/span corruption and machine translation \citep{shazeer2017outrageously,lepikhin2020gshard,kim2021scalable,fedus2021switch}.
In contrast, discrepancies between strong pre-training quality and poor fine-tuning quality for sparse models have been observed in \cite{fedus2021switch,narang2021transformer,artetxe2021efficient}, but we expect advances in regularization techniques to continue to improve downstream quality.

\textbf{Towards better routing algorithms.}
BASE layers \citep{lewis2021base} recasts token routing as a linear assignment problem -- removing the need for load balancing auxiliary losses. This work also demonstrated the efficacy of a single expert layer. 
\cite{clark2022unified} studies in depth the scaling properties of a few different routing algorithms and propose their own variant of BASE layers that uses an optimal transport formulation.
\cite{yang2021m6t} introduces the M6-T architecture and expert prototyping which splits experts into different groups and applies $k$ top-1 routing procedures (contrasting with the top-$k$ routing commonly used elsewhere).
\cite{hazimeh2021dselectk} proposed a continuously differentiable sparse gate with demonstrated improvements over vanilla top-$k$ gating.
Other work \citep{bengio2016conditional} considered casting the routing selection as a reinforcement learning problem.
More radical versions remove learning the routing entirely.
Hash layers \citep{roller2021hash} shows \emph{random} fixed routing (per hash functions) led to competitive performance with learned routing. 
\cite{zuo2021taming} also proposed an algorithm which randomly selects experts during training and inference and found gains of 2 BLEU points over Switch Transformers and competitive scores with the larger models of \cite{kim2021scalable}. 
Finally, \cite{fan2021beyond} designs an architecture with explicit language-specific sublayers (rather than allowing arbitrary routing as done in \cite{lepikhin2020gshard}) to yield gains of +1 BLEU.

\textbf{Sparse expert models in other modalities.}
MoE and sparse experts model have also advanced results in modalities aside from language.
\cite{riquelme2021scaling} designed a 15B parameter V-MoE to match state-of-the-art ImageNet \citep{deng2009imagenet} models with fewer computational resources.
\cite{lou2021sparsemlp} similarly showed a benefit over dense vision models by using MoE layers across both image patch and channel dimensions. 
Additionally, Automatic Speech Recognition has been improved by the SpeechMoE variants \citep{you2021speechmoe,you2021speechmoe2}.
\cite{kumatani2021building} reduced word error rates using MoE models in Sequence-to-Sequence Transformer and Transformer Transducer.

\textbf{Improving deployment of sparse models.}
Initial expert designs (including this work) route each token separately to experts at that layer.
One issue is that these type of architectures may be burdensome to serve since it requires sufficient memory for storing the parameters.
Distillation was shown in \cite{fedus2021switch} to be moderately effective, but recent approaches modified the routing to instead route full sentences or tasks \citep{kudugunta2021beyond,zuo2021taming} which then permits extraction of sub-networks at time of serving (e.g.\@ deploy only the network associated with the new task).
As an alternative to distillation, \cite{kim2021scalable} considers directly pruning away experts not essential to the task of interest.

\textbf{Multitask learning with MoE.}
We conclude our tour of recent MoE research with successes in multitask settings.
\cite{ma2018modeling} recommended using a separate gating or router network for each task, an idea that may soon be revisited for Transformer architectures.
Finally, \cite{gururangan2021demix} recommends even greater modularity of language models and conditionally activates experts based on the domain/task label or by an inferred label.

\section{Discussion} \label{sec:discussion}
While this work is on sparse models, these models intersect with many other interesting topics in machine learning such as adaptive computation, low-precision training, scaling principles, and neural network architecture advances.
Our discussion therefore covers a broader range of topics surfaced during this research.

\paragraph{Unpredictable dynamics when pre-training on multilingual data.} We often observe that the same model pre-trained on multilingual data will yield smaller pre-training speed-ups and be more unstable. One hypothesis is that this is due to the variance of sequences per group across batches. As a reminder, we encourage tokens \emph{in a group} to be load-balanced. There are usually only 2-8 sequences per group (higher becomes expensive) where each sequence is written in a single language. Therefore, at most 2-8 languages must be balanced across experts -- even when training with over 100 languages. This leads to high variance across groups and batches, resulting in chaotic and unpredictable routing. In a follow-up experiment (just highlighted for brevity), we pre-trained on a mixture of English C4 plus a small fraction of a fine-tuning task which similarly resulted in an unstable model.

\paragraph{The robustness of sparse models.} Despite a paper focused on the details of sparse model-particulars, zooming out we find them to be robust to a wide set of hyperparameters and architectural changes. 
Sparse models obtain great performance under a variety of routing algorithms, dropping high fractions of tokens, and different hyperparameters. 
While we did point out the importance of tuning the batch size and learning rate for fine-tuning, our intuition, in-line with \cite{kaplan2020scaling}, is that the real winner is scale.
For instance, Table \ref{tab:top2_vs_top1} shows larger gains to be had by simply increasing the capacity factor (i.e.\@ FLOPs) rather than by more sophisticated routing (i.e.\@ algorithms).

\paragraph{Adaptive computation.} Sparse models are a subclass of adaptive computation models since each input gets different computation applied to it. In sparse models a token is routed to the expert(s) of its choosing. When capacity factors are less than one, the model learns to not apply computation to certain tokens. This has shown promise in computer vision \citep{riquelme2021scaling} and our language experiments (Appendix~\ref{sec:switch_max}). We envision future models expanding this through heterogeneous experts (e.g. each expert applies differing computation). Intuitively, different input examples will likely require different amounts of processing depending on difficulty. Future models in this direction will be efficiently enabled through emerging computing infrastructures \citep{dean2021pathways}.

\paragraph{Generalizing findings from small to large scale.} A key issue we faced throughout our work was identifying small scale models and training setups that reflect larger scale experiments. This was evident in our stability studies in Section~\ref{sec:stability} where experiments had to be run with XL sized models to surface relevant dynamics.
For our architecture and routing algorithm experiments, we often find improvements vanish, or even reverse, when models are trained for longer or made larger.  
As one example, the top-$n$ findings of \cite{fedus2021switch} were reversed in our 8x larger-scale experiments presented here, which revealed small boosts of top-$(n+1)$ routing over top-$n$ routing (see Table \ref{tab:top2_vs_top1}).

\paragraph{Training models with even lower precision.} The best method we found to stabilize our models without hurting (and sometimes improving) quality was the router z-loss. This is an auxiliary loss that encourages the model logits to have values smaller in absolute magnitude. Given the max range of numbers \texttt{float32} and \texttt{bfloat16} can support ($\sim3e^{38}$), this leads us to believe most of this range is not needed, and compressing it actually might improve model training dynamics. Therefore, future precision formats might take into account more compressed exponential ranges to train certain classes of models.

\paragraph{Designing new operations with more multiplicative interactions.} Section~\ref{sec:stability_mult} shows that operations with more multiplicative interactions than additions, or those that don't accumulate over many numbers, improve model performance. We test this further by injecting more multiplicative interactions into expert layers which speedup pre-training by 4\% without any change to step-time (Appendix~\ref{appendix: architecture}). We think this hints at promising architectural improvements for models and could be a good design principle. Recently depthwise convolutions, which only accumulate 3-5 elements, have also been shown to greatly improve Transformer performance ~\citep{so2021primer}. These operations are especially exciting as elementwise multiplications typically do not introduce any communication overhead when using model parallelism (which makes operations like depthwise convolutions and our multiplicative interactions very efficient). While we did note these methods to increase model instabilities in Section~\ref{sec:stability_mult}, using the router z-loss in our models prevented any further instabilities.

\paragraph{Constrain activations to alleviate other undesirable model scaling dynamics.} We observed two additional sources of training instability.  \textbf{(1)} Encoder-decoder models are more unstable than decoder only models (for fixed amount of FLOPs). Encoder-decoder models have a higher ratio of attention layers (e.g. more exponential functions) due to having both self-attention and enc-dec attention layers for each FFN on the decoder. \textbf{(2)} Deeper models are more unstable than shallower models for a fixed amount of FLOPs. Deeper models also introduce more exponential functions through additional attention layers. We hypothesize that a contributing factor to both of these observations is simply the increased number of exponential functions found in the network. Future work could look at resolving these training dynamics by adding z-loss penalties to the attention softmaxes for non-sparse models, especially since we observed adding them didn't change model quality.

\paragraph{Dense and sparse models depend differently on hyperparameters.} 
Our fine-tuning analysis in Section~\ref{sec:finetuning_sensitivity} shows optimal fine-tuning hyperparameters differ significantly between dense and sparse models. 
In certain settings, fine-tuning hyperparamters that worked well for the dense model masked any improvements from the sparse model (despite large pre-training speedups). 
For new model classes, we recommend researchers and practitioners to extensively test key hyperparameters before prematurely abandoning a method.

\section{Conclusion}
We temper the over-exuberance for scale in \citet{fedus2021switch} by showing how a model with 1/5th the size, but with a better balance of computation (FLOPs) to parameters -- is a more effective sparse learner. Furthermore, this improves the usability of sparse models since it can be deployed with less memory overhead.
Using our sparse model variant, we achieve SOTA across a wide range of the most competitive public benchmarks.
We hope this work shows the power of model sparsity and accelerates the adoption of such models.

\section*{Acknowledgements}
We would like to thank Alex Passos, Ekin Cubuk, Margaret Li, Noah Constant, Oriol Vinyals, Basil Mustafa, Joan Puigcerver, 
Diego de Las Casas, Mike Lewis, and Ryan Sepassi for detailed comments and feedback on early versions of the draft. We also thank the Google Brain Team for useful discussions throughout the course of this work. 

\clearpage
\bibliographystyle{plainnat}
\bibliography{iclr2020_conference}
\appendix
\section{Token Load Balance Description}\label{appendix:load_balance}
The auxiliary load balancing loss from \cite{shazeer2017outrageously} is also used to here to balance tokens across experts. 
Assume we have $N$ experts indexed by $i=1$ to $N$ and a batch $\mathcal{B}$ with $T$ tokens.
The auxiliary loss is computed as the scaled dot-product between vectors f and P,
\begin{equation}  \label{eq:total_loss}
\text{loss} = \alpha \cdot N \cdot \sum_{i=1}^{N} f_i \cdot P_i
\end{equation}
where $f_i$ is the fraction of tokens dispatched to expert $i$,
\begin{equation} \label{eq:token_sum}
f_i = \frac{1}{T}\sum_{x \in \mathcal{B}} \mathbbm{1} \{\text{argmax}\: p(x), i\} 
\end{equation}
and $P_i$ is the fraction of the router probability allocated for expert $i$, 
\footnote[2]{A potential source of confusion: $p_i(x)$ is the probability of routing token $x$ to expert $i$. $P_i$ is the probability fraction to expert $i$ across \emph{all tokens} in the batch $\mathcal{B}$.}
\begin{equation} \label{eq:prob_sum}
P_i = \frac{1}{T}\sum_{x \in \mathcal{B}} p_i(x)
\end{equation}
Since we seek uniform routing of the batch of tokens across the $N$ experts, we desire both vectors to have values of $1/N$.
The auxiliary loss of Equation \ref{eq:total_loss} encourages uniform routing since it is minimized under a uniform distribution. The objective can also be differentiated as the $P$-vector is differentiable, but the $f$-vector is not.
The final loss is multiplied by expert count $N$ to keep the loss constant as the number of experts varies since under uniform routing $\sum_{1}^N (f_i\cdot P_i) = \sum_{1}^N (\frac{1}{N}\cdot \frac{1}{N}) = \frac{1}{N}$.
Finally, a hyperparameter $\alpha$ is a multiplicative coefficient for these auxiliary losses; throughout this work we use an $\alpha=10^{-2}$ which was sufficiently large to ensure load balancing while small enough to not to overwhelm the primary cross-entropy objective.

\section{Router Z-Loss Training Dynamics}
\label{app: router_z_loss}
\begin{figure}[ht!]
    \centering
    \includegraphics[width=0.49\columnwidth]{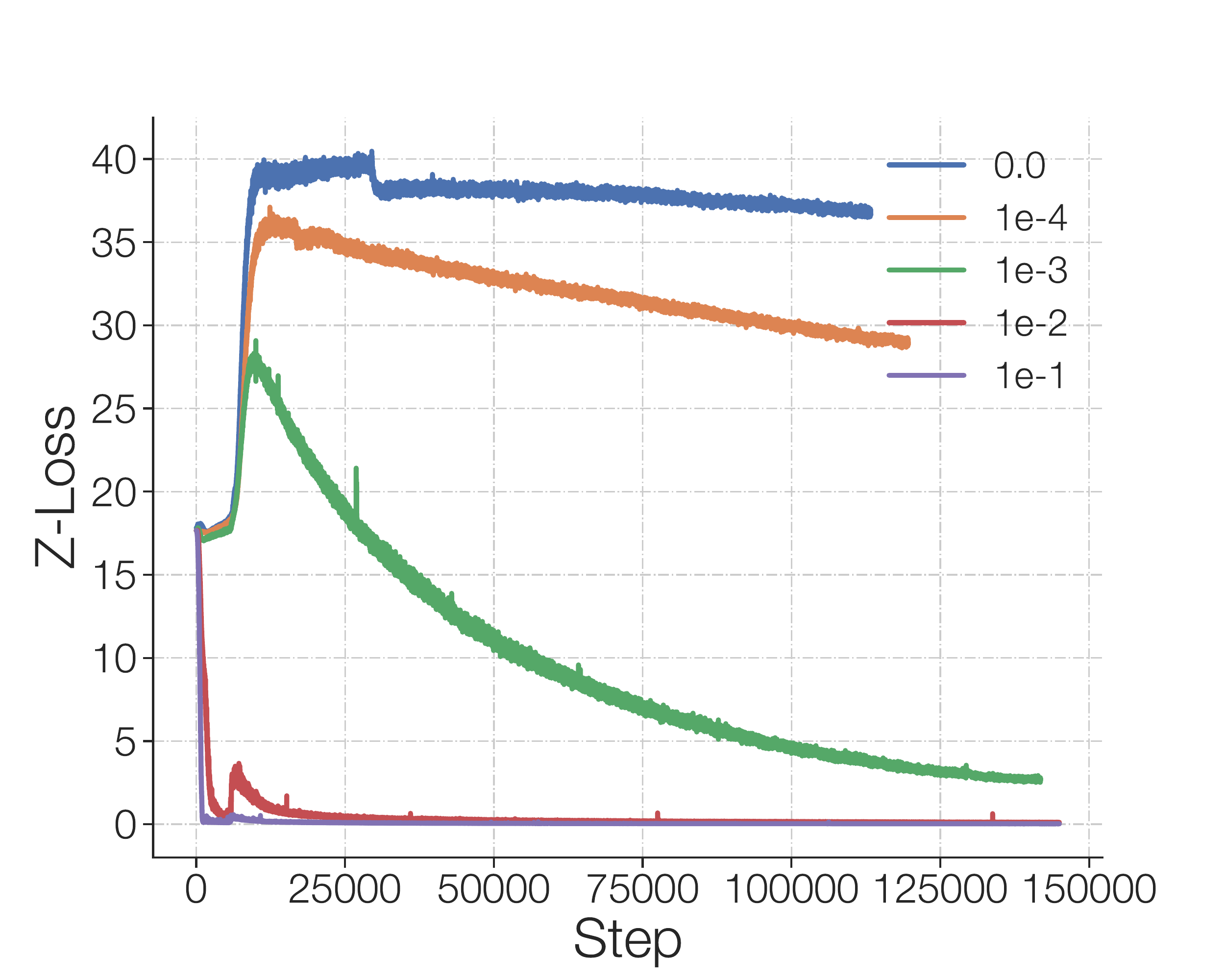}
    \includegraphics[width=0.49\columnwidth]{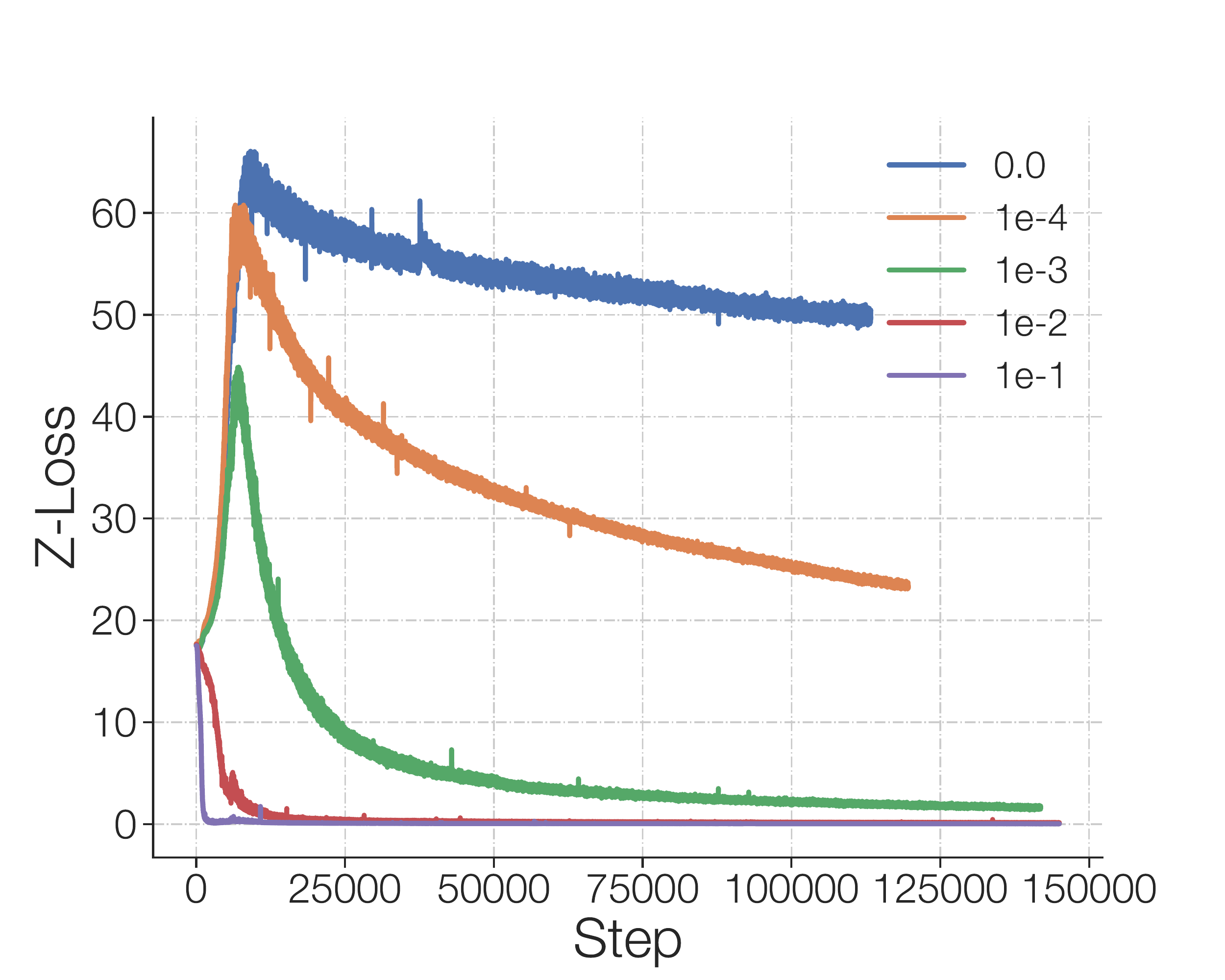}
    \caption{\textbf{Sweeping loss coefficient ($c_z$) for Router Z-Loss.} We plot the router z-losses over the course of pre-training without router z-loss (blue) and with increasing values of $c_z$ (we selected coefficient associated with green curve for all later experiments). With values of \texttt{1e-2}, or larger, the z-loss shrinks near to zero. The left plot shows an encoder layer and the right plot shows a decoder layer. 
}
    \label{fig:router_z_loss}
\end{figure}

Figure \ref{fig:router_z_loss} plots the router z-loss from Equation \ref{eqn: z_loss} across a coefficient sweep where the best value of $c_z=0.001$ is plotted in green for the encoder and decoder.

\section{Improved Architectural Modifications}
\label{appendix: architecture}
We consider a few small architecture variations here.
The first modification was adding additional FFN layers (feed-forward network, see Table \ref{tab:terminology} for more details) immediately before or after each MoE layer (referred to as Sparse-Dense).
Table \ref{tab:sparse_dense} reveals the effectiveness of an FFN layer immediately preceding or following each sparse layer and that these extra FFN layers help less when added elsewhere in the network. 
Guaranteeing all tokens have at least one FFN applied to them between each attention layer appears useful.

\begin{table}[H]
    \centering
    \begin{tabular}{l|c|c}
        \toprule
         \textbf{Model} & \textbf{Neg. Log Perp. ($\uparrow$)} & \textbf{$\Delta$}  \\
         \midrule
         Dense model (baseline) & -1.474 & -  \\
         Dense model w/ extra FFN layers & -1.452 & 0.022  \\
         \midrule
         Sparse model (baseline) & -1.383 & - \\
         Sparse model w/ extra FFN layer \emph{after} each sparse layer & \textbf{-1.369} & 0.014 \\
         Sparse model w/ extra FFN layer \emph{before} each sparse layer & \textbf{-1.369} & 0.014 \\
         Sparse model w/ extra FNN layers placed randomly in the network & -1.376 & 0.007 \\
         \bottomrule
    \end{tabular}
    \caption{\textbf{A dense FFN immediately before or after each sparse layer improves quality}. Inserting an extra dense FFN immediately before or after each sparse layer improves quality 2x as much as placing the dense layers (randomly) elsewhere in the network. All of the non-baseline models have the same amount of FFN layers added for fair comparisons.
    Note that improving perplexity becomes harder as the model gets better.
    }
    \label{tab:sparse_dense}
\end{table}

Second, we introduce an additional bias in the expert layers. 
All our models use the GELU-Linear FFN \citep{shazeer2020glu}, rather than the ReLU FFN:
\begin{equation*}
\begin{split}
\text{FFN}_{\text{ReLU}}(x) = (\text{ReLU}(xW_{1}))W_2 \\
\text{FFN}_{\text{GEGLU}}(x) = (\text{GELU}(xW_{11}) \odot xW_{12} )W_2 \\
\end{split}
\end{equation*}

The additive bias is a learned weight ($B$) added after the first matrix multiplication in the FFN layer of shape $[batch, d_{ff}]$. 
The multiplicative bias (also referred to as a scale parameter) is a learned weight of the same shape, but does an elementwise multiplication.
We initialize the additive bias to zeros and the multiplicative bias to ones.
\begin{equation*}
\begin{split}
\text{FFN}_{\text{GEGLU}} + \text{Add Bias}(x) = [(\text{GELU}(xW_{11}) \odot xW_{12} ) + B]W_2 \\
\text{FFN}_{\text{GEGLU}} + \text{Mult Bias}(x) = [(\text{GELU}(xW_{11}) \odot xW_{12} ) \odot B]W_2 \\
\end{split}
\end{equation*}

Table~\ref{tab:multiplicative_interactions} shows the results of our different methods. 
Both the additive and multiplicative biases are essentially free: cheap to compute, adds few new parameters, and incurs no additional communication costs with model and expert parallelism. When using our router z-loss from Section \ref{sec:stability_mult}, we observe no instabilities from the multiplicative bias.
We do see that the multiplicative interactions improve performance, achieving a 4\% speedup in convergence time over our strong sparse baseline. 
This hints that a promising avenue for future architectural research is finding new ways of adding more multiplicative interactions into networks.

\begin{table}[H]
    \centering
    \begin{tabular}{c|cc}
        \toprule
        \textbf{Model} & \textbf{Neg. Log. Perp. ($\uparrow$)} & $\Delta$  \\
        \midrule
        Dense Baseline & -1.474 & - \\
        Sparse Baseline & -1.369 & - \\
        \midrule
        Sparse $+$ Additive Bias & -1.371 & -0.002 \\ 
        Sparse $+$ Multiplicative Bias & \textbf{-1.361} & 0.008 \\ 
        \bottomrule
    \end{tabular}
    \caption{\textbf{More multiplicative interactions improve sparse model quality.} Both the additive and the multiplicative bias add virtually no parameters or compute.}
    \label{tab:multiplicative_interactions}
\end{table}

Finally, motivated by the work of \cite{roller2021hash}, we explored similar methods, but did not find improvements in our setting. 
We tried routing using the word embedding exclusively, as well as an additional input to the layer embedding for routing decisions. 
We toggled stopping the gradient through the word embedding or allowing it to have gradients propagated from the router.
Using only the word embedding hurt quality, while using it in addition to the normal layer hidden activation was initially positive, but after pre-training for 50B+ tokens on models of scale 1B+ dense parameters it had a neutral effect. 
Appendix~\ref{sec:negative} has further details on the experiments with negative results.

\section{Batch Prioritized Routing for Lower Capacity Factors}
\label{sec:switch_max}
Surprisingly, top-1 and top-2 routing work well with CF less than 1.0 despite token routing being done in a left to right order over the sequence.
If $N$ tokens are sent to an expert with only $M$ spaces then $N > M$ tokens will dropped.
The ordering of the dropping is important: we drop tokens going left to right (e.g.\@ tokens earlier in the sentence will be routed first over the end tokens).
This is done to avoid the model cheating. 
If we dropped tokens in another ordering, the model gets information on what tokens are occurring later in the sequence based on if tokens are being dropped or not.

Batch Prioritized Routing (BPR) from \cite{riquelme2021scaling} was introduced in Vision Transformers \citep{dosovitskiy2020image} for image classification.
Our work explores BPR with top-1 routing in the context of language modeling.
BPR aims to have a global view of all tokens to determine which tokens should be dropped instead of the left-to-right ordering.
The algorithm works by looking at all N tokens getting sent to Expert $i$ and then only routing the $M$ ones with the highest probabilities from the router.
Table~\ref{tab:switch_max} shows that BPR top-1 routing improves performance over top-2 routing, especially when capacity factors are less than 1.0. We leave it to future work to try top-$n$ BPR routing, which will hopefully yield larger improvments for higher capacity factors.

Importantly, BPR routing can only be done on the encoder side of the encoder-decoder model.
On the encoder side there are not autoregressive predictions and all tokens can see each other.
If you use BPR on the decoder, it learns to cheat by using future token information to improve current token predictions.

\begin{table}[H]
    \centering
    \begin{tabular}{c|cc|c}
         \toprule
         \textbf{Algorithm} & \textbf{Train CF} & \textbf{Eval CF} & \textbf{Neg. Log. Perp. ($\uparrow$)} \\
         \midrule
         Dense & --- & --- & -1.474  \\
         Dense-L & --- & --- & -1.384  \\
         \midrule
         \rowcolor{blue!15}
         BPR Top-1 & 0.5 & 0.5 & -1.433 \\
         \rowcolor{blue!15}
         BPR Top-1 & 0.5 & 2.0 & -1.416 \\
         \midrule
         \rowcolor{gray!15}
         Top-1 & 0.75 & 0.75 & -1.428 \\
         \rowcolor{gray!15}
         Top-1 & 0.75 & 2.0 & -1.404 \\
         Top-2 & 0.75 & 0.75 & -1.424 \\
         Top-2 & 0.75 & 2.0 & -1.402 \\
        \rowcolor{blue!15}
         BPR Top-1 & 0.75 & 0.75 & -1.409 \\
         \rowcolor{blue!15}
         BPR Top-1 & 0.75 & 2.0 & -1.397 \\
         \midrule
         \rowcolor{gray!15}
         Top-1 & 1.0 & 1.0 & -1.397 \\
         \rowcolor{gray!15}
         Top-1 & 1.0 & 2.0 & -1.384 \\
         Top-2 & 1.0 & 1.0 & -1.392 \\
         Top-2 & 1.0 & 2.0 & -1.378 \\
        \rowcolor{blue!15}
         BPR Top-1 & 1.0 & 1.0 & -1.386 \\
         \rowcolor{blue!15}
         BPR Top-1 & 1.0 & 2.0 & -1.379 \\
         \midrule
         \rowcolor{gray!15}
         Top-1 & 1.25 & 1.25 & -1.378 \\
         \rowcolor{gray!15}
         Top-1 & 1.25 & 2.0 & -1.373 \\
         Top-2 & 1.25 & 1.25 & -1.375 \\
         Top-2 & 1.25 & 2.0 & -1.369 \\
         \rowcolor{blue!15}
         BPR Top-1 & 1.25 & 1.25 & -1.376 \\
         \rowcolor{blue!15}
         BPR Top-1 & 1.25 & 2.0 & -1.375 \\
         \bottomrule
    \end{tabular}
    \caption{\textbf{Batch Prioritized Top-1 Routing (BPR) performance.} BPR top-1 routing improves quality when capacity factors are $\leq$ 1. However, once the capacity factor reaches 1.25, the improvements greatly diminish and it underperforms top-2 routing. Future work can try BPR with top-2 routing, which should hopefully further improve the performance.}
    \label{tab:switch_max}
\end{table}

\section{Pre-Training Dataset Details}\label{app: pretrain_data}
The pre-training dataset used to train our Sparse 32B model is a mix of C4 \citep{raffel2019exploring} and the dataset introduced in GLaM \citep{du2021glam}. 

\begin{table}[h!]
\begin{center}
\small
\begin{tabular}{lccc}
\toprule
\textbf{Dataset} & \textbf{Tokens (B)} & \textbf{Weight in Mixture} \\
\midrule
Filtered C4 & 183 & 0.17 \\
Filtered Webpages & 143 & 0.34 \\
Wikipedia & 3 & 0.05\\
Conversations  & 174 & 0.23 \\
Forums & 247 & 0.02 \\
Books & 390 & 0.17 \\
News & 650 & 0.02 \\
\bottomrule
\end{tabular}
\caption{\textbf{Data and mixture weights in the training set}. We sample from different dataset sources with probability proportional to ``weight in mixture''. The number of tokens listed are in billions~(B). For more details on the C4 corpus see \cite{raffel2019exploring} and for the other datasets see \cite{du2021glam}.}
\label{tab:data}
\end{center}
\end{table}

\clearpage
\section{Full Fine-tuning Sensitivity Data}
\label{appendix: fine_tuning_protocol}
Table \ref{tab:fine_tuning_protocol} contains the raw data for Figure \ref{fig:dense_vs_expert_scaling} measuring the fine-tuning protocol sensitivity. Dense and Sparse are encoder-decoder models FLOP matched to T5-Large that were pre-trained for 500k steps with a batch size of 1M tokens on the C4 corpus.

\begin{table}[H]
    \centering
    \begin{tabular}{cccc|cc}
        \toprule
        \textbf{Model} & \textbf{Learning Rate} & \textbf{Batch Size} & \textbf{Reset Optimizer Slot Vars} & \textbf{SuperGLUE ($\uparrow$)} \\
        \midrule
        Dense & 1e-3 & 1M & & 84.8 \\
        Dense & 1e-3 & 1M & \checkmark & 84.3 \\
        Dense & 5e-4 & 1M & & 84.8 \\
        Dense & 5e-4 & 1M & \checkmark & 84.2 \\
        Dense & 1e-4 & 1M & & 84.0 \\
        Dense & 1e-4 & 1M & \checkmark & 84.8 \\
        \midrule
        Dense & 1e-3 & 262k & & 84.9 \\
        Dense & 1e-3 & 262k & \checkmark & 83.7 \\
        Dense & 5e-4 & 262k & & 84.9 \\
        Dense & 5e-4 & 262k & \checkmark & 84.0 \\
        Dense & 1e-4 & 262k & & \textbf{85.1} \\
        Dense & 1e-4 & 262k & \checkmark & 85.0 \\
        \midrule
        Dense & 1e-3 & 65k & & 83.7 \\
        Dense & 1e-3 & 65k & \checkmark & 82.5 \\
        Dense & 5e-4 & 65k & & 84.4 \\
        Dense & 5e-4 & 65k & \checkmark & 84.1 \\
        Dense & 1e-4 & 65k & & 84.9 \\
        Dense & 1e-4 & 65k & \checkmark & 84.6 \\
        \midrule
        \midrule
        Sparse & 1e-3 & 1M & & \textbf{86.9} \\
        Sparse & 1e-3 & 1M & \checkmark & 85.9 \\
        Sparse & 5e-4 & 1M & & 86.1 \\
        Sparse & 5e-4 & 1M & \checkmark & 83.5 \\
        Sparse & 1e-4 & 1M & & 84.3 \\
        Sparse & 1e-4 & 1M & \checkmark & 84.3 \\
        \midrule
        Sparse & 1e-3 & 262k & & 86.2 \\
        Sparse & 1e-3 & 262k & \checkmark & 85.2 \\
        Sparse & 5e-4 & 262k & & 85.5 \\
        Sparse & 5e-4 & 262k & \checkmark & 84.8 \\
        Sparse & 1e-4 & 262k & & 85.1 \\
        Sparse & 1e-4 & 262k & \checkmark & 85.5 \\
        \midrule
        Sparse & 1e-3 & 65k & & 85.8 \\
        Sparse & 1e-3 & 65k & \checkmark & 85.5 \\
        Sparse & 5e-4 & 65k & & 86.5 \\
        Sparse & 5e-4 & 65k & \checkmark & 85.1 \\
        Sparse & 1e-4 & 65k & & 85.6 \\
        Sparse & 1e-4 & 65k & \checkmark & 84.5 \\
        \bottomrule
    \end{tabular}
    \caption{\textbf{Fine-tuning protocol sensitivity.} We vary the batch size, learning rate and whether we reset the optimizer slot variables for both dense and sparse models. Resetting the optimizer state during fine-tuning hurts performance. We observe a difference in optimal batch size and learning rate for sparse vs. dense models. Certain hyperparameter fine-tuning settings make the sparse and dense models perform almost exactly the same, showing the importance of correctly tuning the hyperparameters.}
    \label{tab:fine_tuning_protocol}
\end{table}

\section{Optimally Setting the Routing Threshold}

\begin{tcolorbox}[enhanced,attach boxed title to top center={yshift=-3mm,yshifttext=-1mm}, title=\textbf{Top-$n$ Routing Algorithm}, colback=white, colframe=white!75!blue, coltitle=black, colbacktitle=white]
    \begin{enumerate}[leftmargin=*,label=\textbf{\arabic*.}]
    \item Route each token $x$ to the expert with the highest router probability ($\text{gate}_1(x)$).
    \item Normalize the top-$n$ expert router scores for each token $x$, so $\text{gate}_i = \frac{\text{gate}_i(x)}{\sum_{i=1}^n \text{gate}_i(x)}$.
    \item Route the token to the other $n$-1 experts (indexed by $i$) with probability $\text{min}(1.0, \frac{\text{gate}_i(x)}{\text{threshold}})$. Threshold is a predefined hyperparameter that is typically set to 0.2.
    \end{enumerate}
\end{tcolorbox}

We describe the MoE hyperparameters and how they should change as the routing algorithm changes. The MoE top-2 routing algorithm \citep{shazeer2017outrageously,shazeer2018mesh,lepikhin2020gshard} works as follows: first the router finds the expert that is assigned the higher router score ($\text{gate}_1$) and always sends the token to that expert. 
The token is also sent to its second highest expert with probability $\text{min}(1.0, \text{gate}_2 / \text{threshold})$.
The $\text{threshold}$ is a hyperparameter that is typically set to 0.2, and $\text{gate}_2$ is the token's router probability for the second highest expert. Note that $\text{gate}_1$ and $\text{gate}_2$ get normalized by the sum of their two scores, so they sum to one.

We trivially extend the top-2 algorithm to work for top-$n$ routing here. Take the scores of the top-$n$ experts per token and sum them, then renormalize each expert router score based on that sum. If the specific renormalized expert score has a higher value than the threshold (e.g.\@ 0.2), then the token will be routed, otherwise it will be routed with probability $\frac{\text{score}}{\text{threshold}}$. At a high level this  only routes the token to the next $n$-1 experts if their scores are not too much lower than the highest scored expert.

For top-3 routing vs top-2, the sum that the expert scores are normalized by is larger, therefore we experimented with \emph{decreasing} the threshold. Our experimental results are shown in Table~\ref{tab:moe_threshold}. Interestingly, we do observe the top-3 routing to slightly benefit from the lower threshold, while the opposite is true for top-2 routing.

We also experimented with an absolute threshold policy instead of a relative one. This is where the next $n$-1 tokens will be routed only if their router score is great than some pre-defined value (e.g. 0.2). We found it can achieve as good of performance if the threshold value is tuned.

\begin{table}[H]
    \centering
    \begin{tabular}{c|cc|c}
         \toprule
         \textbf{Algorithm} & \textbf{Train CF} & \textbf{Threshold} & \textbf{Neg. Log. Perp. ($\uparrow$)} \\
         \midrule
         Dense & --- & --- & -1.474 \\
         Dense-L & --- & --- & -1.384 \\
         \midrule
         Top-2 & 3.0 & 0.2 & -1.354 \\
         Top-2 & 3.0 & 0.05 & -1.356 \\
         Top-3 & 3.0 & 0.2 & \textbf{-1.351} \\
         Top-3 & 3.0 & 0.05 & \textbf{-1.349} \\
         \bottomrule
    \end{tabular}
    \caption{\textbf{Performance of top-2 and top-3 routing with different thresholds.} Top-3 routing does slightly better with lower thresholds than top-2 routing.}
    \label{tab:moe_threshold}
\end{table}

\section{Mesh Layout for Data, Model and Expert Parallelism with Few Experts}
\label{lab:mesh_explanation}
\begin{figure}[ht!]
    \centering
    \includegraphics[width=0.5\columnwidth]{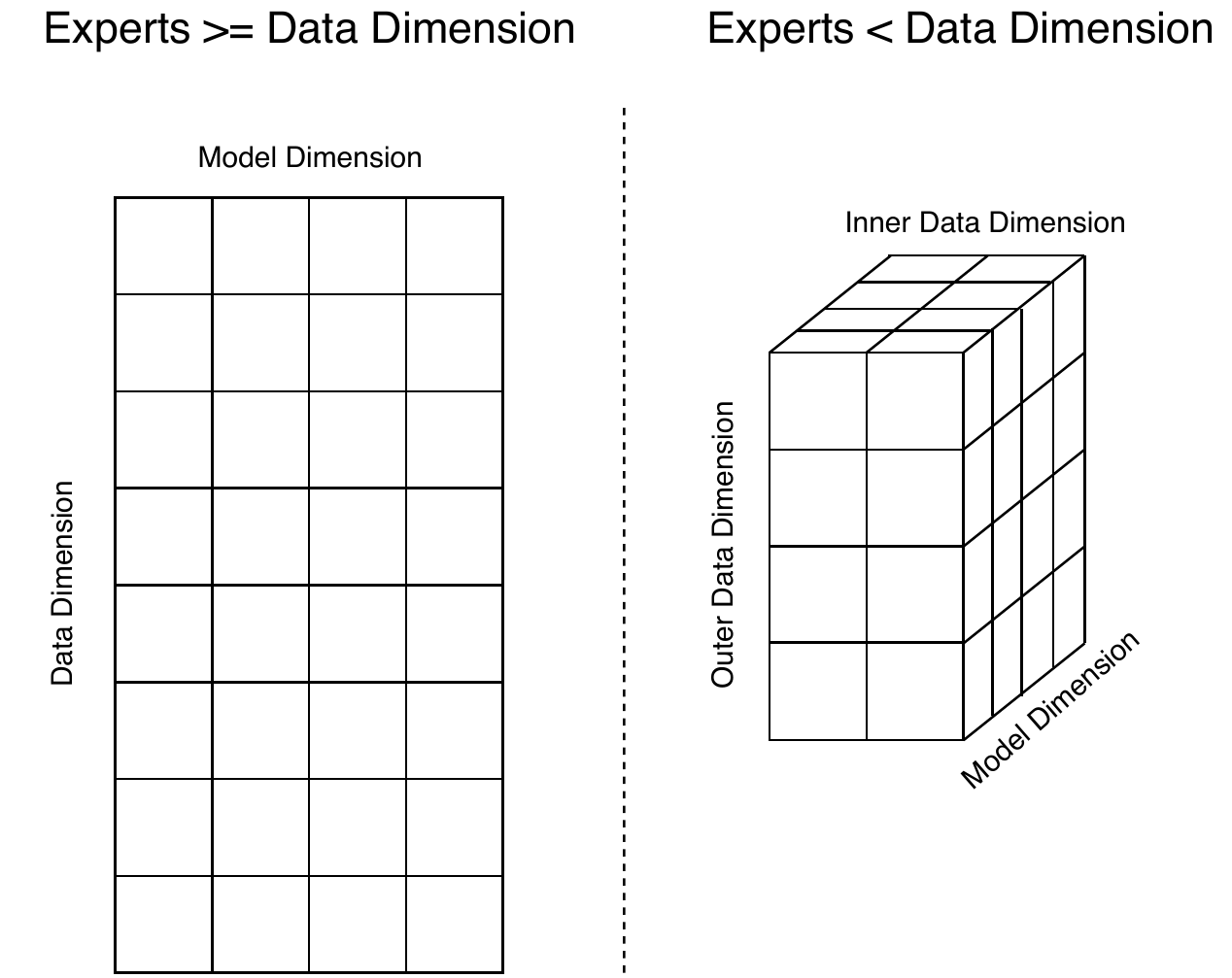}
    \caption{\textbf{Data and model parallelism meshes used for distributing models.} In this example there are a total of 32 processors (e.g. $n=32$). \textbf{(Left)} A valid 2D mesh if the number of experts is greater than or equal to the data parallelism dimension. The data dimension has 8 rows ($d$) and the model dimension has 4 columns ($m$). \textbf{(Right)} A valid 3D mesh when we have fewer experts than the data parallelism dimension. The batch dimension is factorized into two new dimensions: inner data and outer data dimensions. Now we have 1 expert per inner data dimension ($i$). The 8 data rows in the left figure become 4 in the outer batch ($o$) and 2 in the inner batch ($i$) with 2 experts instead of 8.
}
    \label{fig:mesh_data_model}
\end{figure}
We use data and model parallelism partitioning with Mesh-Tensorflow \citep{shazeer2018mesh}.
The partitioning strategy works by first forming a logical 2D mesh of size $d$ x $m$, with the rows corresponding to the data dimension ($d$) and the columns as the model dimension ($m$) and the product equal to the total number of cores, $n$ = $d$ x $m$. 
This mesh is only an abstraction.
Each logical core must be mapped to a physical core, which is optimized through performance tuning.

As a refresher, each row in the mesh will have its own unique slice of the data and each column will have a unique slice of the model weights.
The final gradient \texttt{allreduce} communication occurs across each individual column.
The model parallelism \texttt{allreduce} communications occur across each row in the mesh. 
One constraint from this approach is that the number of rows must evenly divide the number of data sequences and the number of columns must evenly divide the model dimensions being partitioned.

But if we have \emph{fewer} than $d$ experts then this layout will not work.
To allow for fewer experts than data parallelism rows in our mesh, we factorize the data dimension into two new dimensions: inner ($i$) and outer ($o$) where $i$ x $o$ = $d$ and the number of experts equals $i$. 
This transforms the logical 2D mesh of shape $d$ x $m$ into a 3D mesh of shape $o$ x $i$ x $m$. 
See Figure \ref{fig:mesh_data_model} for a visualization of both meshes \footnote{See Mesh Tensorflow for more details on the inner and outer batch: \url{https://github.com/tensorflow/mesh/blob/master/mesh_tensorflow/transformer/moe.py}}.

\section{Note on Communication Costs for Distributed Models}
\label{sec:note_communication}
Communication operations (\texttt{allreduce} and \texttt{all2all}) can significantly impact sparse model training throughput (see Table~\ref{tab:terminology} for a description of the communication operations).
\texttt{allreduce} calls are executed along model and batch dimensions, typically dominated by the model dimension \texttt{allreduce} calls that sum results of partial matrix multiplication operations from the workers.
These calls are needed when matrix multiplications are partitioned across multiple cores (e.g.\@ model parallelism).
The gradient summation \texttt{allreduce} calls can be amortized away by training models with larger batch sizes since the gradient accumulation \texttt{allreduce} communication cost is independent of the batch size.
To alleviate the memory issues of larger batch sizes, microbatches can be used.
Microbatches do this by splitting the batch into $n$ evenly divisible chunks and computing gradients on each sequentially, then summing.

To increase the \texttt{allreduce} throughput, more workers may need to be assigned to the model dimension (instead of batch dimension).
However, increasing the number of workers may reduce compute per worker resulting in higher communication overheads that cancel some of the gains from higher communication throughput from \texttt{allreduce}.
For the results in this paper, first we explored various model partitioning strategies.
Next the shapes of the pre-training jobs were allocated based on performance benchmarking which showed the lowest cumulative communication overheads in \texttt{allreduce} and \texttt{all2all}.

\section{Negative Results}
\label{sec:negative}
We conclude with some ideas that yielded negative results in our setting.

\paragraph{Adding information if tokens were dropped to the router.} We experimented with having the expert layer have information of whether the token was routed or dropped in the previous expert layers. We implemented this through counting the number of times a token was routed in all previous expert layers,  having embeddings for each possible value and then adding this to the router embedding. We found that this made no difference in performance.

\paragraph{Adding explicit expert positional information.} We experimented with adding explicit positional information into the outputs of the expert layer. We wanted to see if it either improved performance or sped up convergence during the beginning of training when expert layers were drastically changing. We did this through adding an embedding corresponding to what expert each token was sent (including an embedding if the token was dropped), but this did not improve performance.

\paragraph{Adding pre-training noise to fix pre-training and fine-tuning discrepancies.} To help fix the pre-training perplexity and fine-tuning gap we tried pre-training the sparse models with a variety of different types of noise. The goal was to help pre-training match the fine-tuning conditions where dropout is used and more tokens can be dropped. Some of the noise types we tried adding during pre-training were dropout, dropping out full experts for a batch of tokens, and adding an entropy maximization auxiliary loss to the router. Unfortunately, all of the methods either hurt the pre-training quality too much or didn't end up helping the fine-tuning.

\paragraph{Load balancing in top-n routing over lower n-1 experts.} In the standard top-$n$ MoE formalization there is only loading balancing over the top expert a token is sent to. We experimented with adding an auxiliary load balancing term to the other $n-1$ experts in top-$n$ routing, but found this to provide minimal benefits.

\paragraph{Mixing pre-training and fine-tuning data to prevent overfitting.} To help combat the overfitting of sparse models during fine-tuning, we tried mixing in pre-training span corruption data at varying amounts (e.g.\@ 1\%, 5\%, 25\%, ...) during fine-tuning. This ended up not helping the fine-tuning performance, but did increase the training loss.

\end{document}